\documentclass{article}

\PassOptionsToPackage{numbers, compress}{natbib}

\usepackage[preprint]{neurips_2023}

\usepackage[utf8]{inputenc} 
\usepackage[T1]{fontenc}    
\usepackage{url}            
\usepackage{booktabs}       
\usepackage{amsfonts}       
\usepackage{nicefrac}       
\usepackage{microtype}      
\usepackage{xspace}
\usepackage{graphicx}
\usepackage{subcaption}
\usepackage{multirow}
\usepackage{multicol}
\usepackage{colortbl}
\usepackage{pifont}
\usepackage{wrapfig}

\usepackage[table]{xcolor}
\usepackage[pagebackref, breaklinks=true, colorlinks, citecolor=citecolor, linkcolor=linkcolor, bookmarks=false]{hyperref}
\definecolor{citecolor}{HTML}{0071BC}
\definecolor{linkcolor}{HTML}{ED1C24}

\def\modelname{VisionLLM\xspace}

\newcommand{\xk}[1]{\textcolor{red}{[xk: #1]}}

\newcommand{\yes}{\ding{51}}
\newcommand{\no}{\ding{55}}

\newlength\savewidth\newcommand\shline{\noalign{\global\savewidth\arrayrulewidth
  \global\arrayrulewidth 1pt}\hline\noalign{\global\arrayrulewidth\savewidth}}

\def\ie{\emph{i.e.}}
\def\eg{\emph{e.g.}}

\def\etal{{\em et al.~}}

\usepackage{pgfplots}
\usepackage{pgfplotstable}
\pgfplotsset{compat=1.8}
\usetikzlibrary{pgfplots.groupplots}
\definecolor{mediumelectricblue}{rgb}{0.01, 0.31, 0.59}
\definecolor{lightsalmon}{rgb}{1.0, 0.63, 0.48}

\makeatletter
\def\blfootnote{\xdef\@thefnmark{}\@footnotetext}
\makeatother

\title{VisionLLM: Large Language Model is also\\ an Open-Ended Decoder for Vision-Centric Tasks}

\author{
	\hspace{-0.25cm}\textbf{Wenhai Wang$^{*1}$, Zhe Chen$^{*2,1}$, Xiaokang Chen$^{*3,1}$, Jiannan Wu$^{*4,1}$}, Xizhou Zhu$^{5,1}$ \\
	\hspace{-0.25cm}\textbf{Gang Zeng$^{3}$, Ping Luo$^{4,1}$, Tong Lu$^{2}$, Jie Zhou$^6$, Yu Qiao$^1$, Jifeng Dai$^{\dagger 6,1}$} \\ 
	\hspace{-0.25cm}$^1$OpenGVLab, Shanghai AI Laboratory \quad $^2$Nanjing University \quad $^3$Peking University \\
	\hspace{-0.25cm}$^4$The University of HongKong \quad $^5$SenseTime Research \quad $^6$Tsinghua University \\\\
	{Code: \url{https://github.com/OpenGVLab/VisionLLM}} \\
	{Demo: \url{https://github.com/OpenGVLab/InternGPT}}
}

\begin{document}

\maketitle

\thispagestyle{empty}

\blfootnote{\noindent $^{*}$Equal contribution. This work is done when Zhe Chen, Xiaokang Chen, and Jiannan Wu are interns at Shanghai AI Laboratory. 
$^{\dagger}$ Corresponding to Jifeng Dai <daijifeng@tsinghua.edu.cn>.}

\begin{abstract}
  Large language models (LLMs) have notably accelerated progress towards artificial general intelligence (AGI), with their impressive zero-shot capacity for user-tailored tasks, endowing them with immense potential across a range of applications. 
  However, in the field of computer vision, despite the availability of numerous powerful vision foundation models (VFMs), they are still restricted to tasks in a pre-defined form, struggling to match the open-ended task capabilities of LLMs.
  In this work, we present an LLM-based framework for vision-centric tasks, termed VisionLLM.
  This framework provides a unified perspective for vision and language tasks by treating images as a foreign language and aligning vision-centric tasks with language tasks that can be flexibly defined and managed using language instructions. An LLM-based decoder can then make appropriate predictions based on these instructions for open-ended tasks.
  Extensive experiments show that the proposed VisionLLM can achieve different levels of task customization through language instructions, from fine-grained object-level to coarse-grained task-level customization, all with good results. It's noteworthy that, with a generalist LLM-based framework, our model can achieve over 60\% mAP on COCO, on par with detection-specific models. We hope this model can set a new baseline for generalist vision and language models. The code and demo shall be released.
\end{abstract}

\section{Introduction}

The emergence of large language models (LLMs) like ChatGPT~\cite{openai2022chatgpt} has revolutionized the landscape of artificial general intelligence (AGI), showcasing their impressive zero-shot capabilities in addressing various natural language processing (NLP) tasks through user-tailored prompts or language instructions. 
Despite these advancements, it's essential to note that the triumph of LLMs does not effortlessly extend to pure vision and vision-language tasks, due to the inherent disparities between modalities and task formats.

\begin{figure}[t]
\hsize=\textwidth
\centering
\begin{subfigure}{0.3\textwidth}
    \centering
    \includegraphics[width=0.95\textwidth]{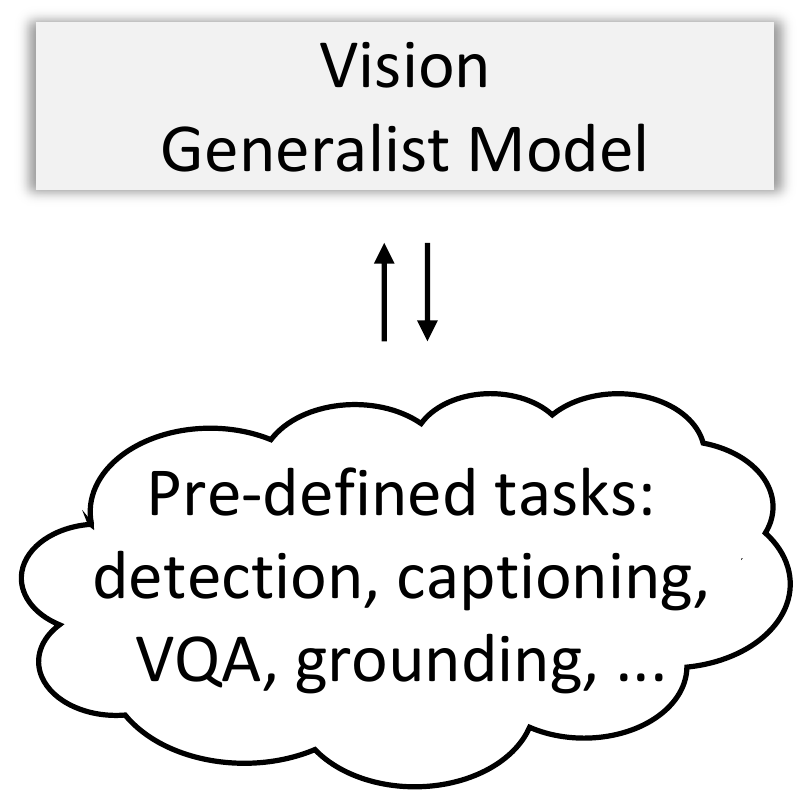}
    \caption{Vision generalist models~\cite{wang2022internimage, wang2022image,zhu2022uni_p} are constrained by the format of pre-defined tasks.}
    \label{fig:1a}
\end{subfigure}    
\hspace{0.2in}
\begin{subfigure}{0.3\textwidth}
     \centering
     \includegraphics[width=0.95\textwidth]{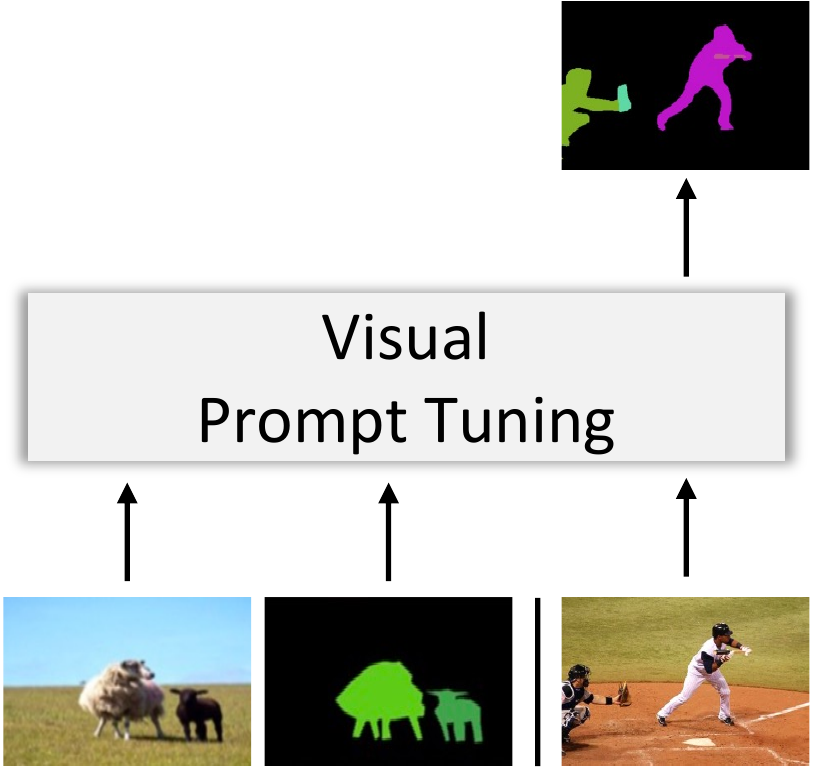}
     \caption{Visual prompt tuning~\cite{jia2022visual,wang2023seggpt,wang2022images} are inconsistent with the format of LLMs.}
     \label{fig:1b}
\end{subfigure}
\hspace{0.2in}
\begin{subfigure}{0.3\textwidth}
     \centering
     \includegraphics[width=0.95\textwidth]{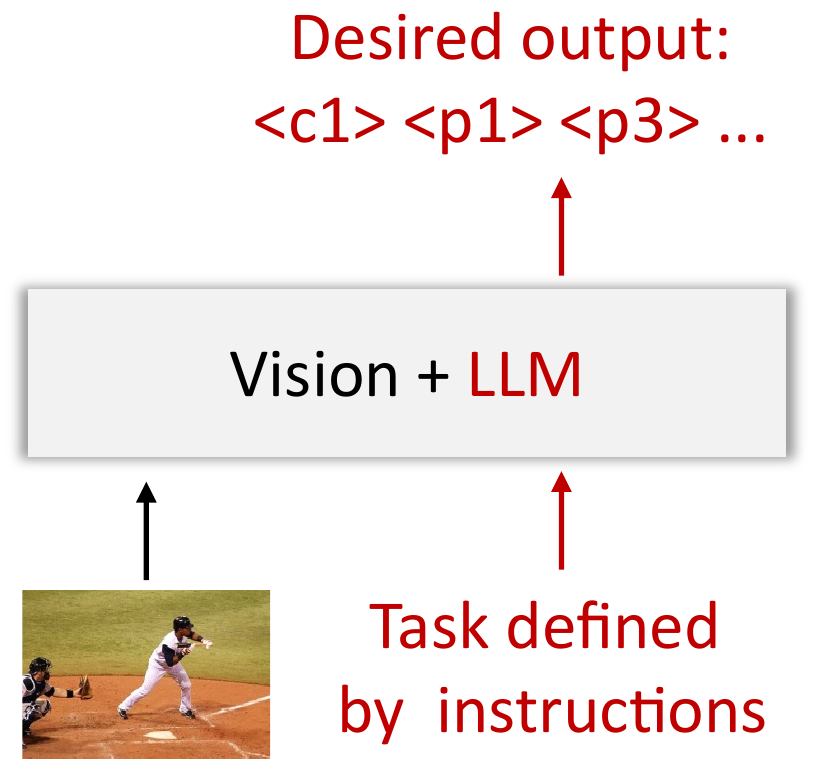}
     \caption{VisionLLM (ours) can \emph{flexibly manage vision-centric tasks using language instructions like LLMs}.}
     \label{fig:1d}
\end{subfigure}
\caption{\textbf{Comparison of our \modelname with popular paradigms.}
Unlike current vision generalist models that depend on pre-defined task formats and visual prompt tuning models that are inconsistent with large language models (LLMs), \modelname leverages the power of LLMs for open-ended vision tasks by using language instructions.
}
\label{fig:cmp}
\end{figure}

The field of computer vision presents a unique set of challenges and paradigms that differ from those of NLP. The traditional paradigm of vision foundation models is pre-training followed by fine-tuning~\cite{wang2022internimage, chen2022vision,su2022towards, wang2022image, fang2022eva,tao2022siamese}, 
which is effective but comes with significant marginal costs when adapting to diverse downstream scenarios.
As shown in Figure \ref{fig:1a}, 
while approaches such as multi-task unification~\cite{radford2018improving,ofa,alayrac2022flamingo,wang2022git,zhu2022uni} have been used to achieve generalist capability, they often struggle to overcome the limitations imposed by pre-defined tasks, resulting in a gap in open-ended task capabilities compared to LLMs.
Recently, visual prompt tuning~\cite{jia2022visual,yao2021cpt,zhang2022neural,zang2022unified,wang2022images} has emerged as a way to flexibly outline some pure vision tasks (see Figure \ref{fig:1b}), such as object detection, instance segmentation, and pose estimation, using visual masking.
However, the format of visual prompts considerably deviates from that of language instructions, making it challenging to directly apply the reasoning abilities and world knowledge of LLMs to vision tasks.
Therefore,
\emph{ 
there is an urgent need for a unified generalist framework that can seamlessly integrate the strengths of LLMs with the specific requirements of vision-centric tasks.
}

In this work, we present VisionLLM, a novel framework that aligns the definitions of vision-centric tasks with the methodologies of LLMs. Leveraging the reasoning and parsing capacities of LLMs, VisionLLM is designed to empower open-ended task capabilities for vision-centric tasks.
Specifically, it comprises three core components: (1) a unified language instruction designed for vision and vision-language tasks, (2) a language-guided image tokenizer, and (3) an LLM-based open-ended task decoder that orchestrates various tasks using language instructions.
With this framework, a wide range of vision-centric tasks can be seamlessly integrated, including object detection, instance segmentation, image captioning, and visual grounding. In addition, the framework also facilitates task customization at different levels of granularity, allowing for the customization of target objects, output formats, task descriptions, etc.

Compared to current popular API-based applications~\cite{wu2023visual, Yang2023MMREACTPC, shen2023hugginggpt,liu2023interngpt,li2023videochat}, our model takes a unified, end-to-end approach to integrate VFMs and LLMs, streamlining and enhancing the overall efficiency of the overall process, and leveraging the strengths and data of both VFMs and LLMs within a single, cohesive system.
Furthermore, our model surpasses the limitations of generalist vision models pre-trained on pre-defined tasks.
VisionLLM can effectively manage vision-centric tasks through language instructions, embodying a flexible and open-ended approach that is not constrained by pre-set tasks.
This versatility makes VisionLLM a robust and powerful generalist model for vision and vision-language tasks, opening up new possibilities for the development of unified generalist models that bridge the domains of vision and language.

In summary, our main contributions are as follows:

(1) We propose VisionLLM, the first framework that leverages the power of LLMs to address vision-centric tasks in an open-ended and customizable manner. By aligning the definitions of vision-centric tasks with LLM methodologies, VisionLLM breaks new ground in enabling the unified modeling of vision and language, opening up possibilities for advancing the field.

(2) We overcome many difficulties when porting LLMs to vision-centric tasks, by designing unified language instruction that matches the format of language models and covers various vision-centric tasks including visual perception.
Correspondingly, we develop a language-guided image tokenizer and an LLM-based task decoder that can handle open-ended tasks according to the given language instructions based on the LLMs' reasoning and parsing capabilities.

\begin{figure}[t]
\hsize=\textwidth
\centering
\begin{subfigure}{0.48\textwidth}
    \centering
    \includegraphics[width=0.96\textwidth]{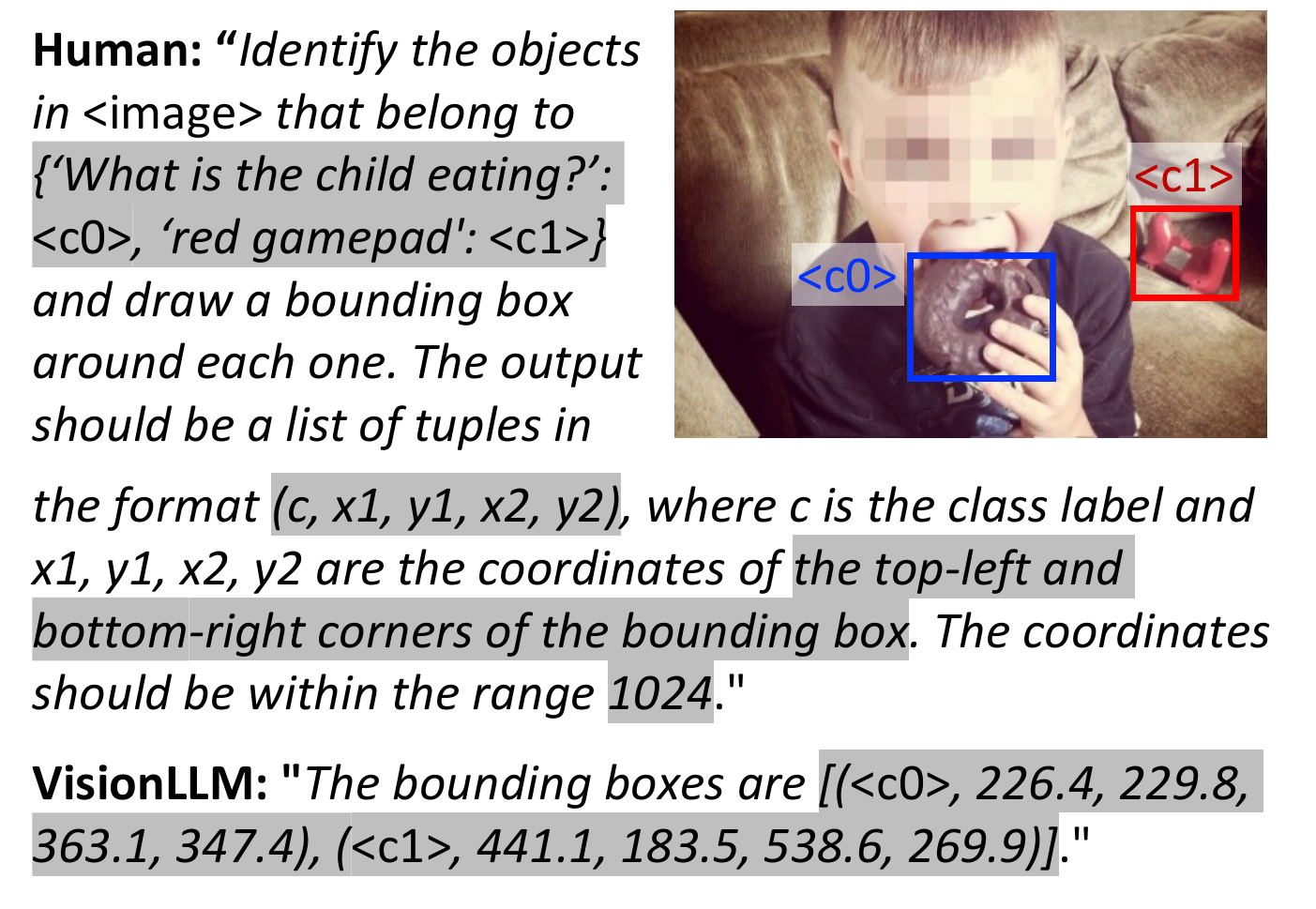}
    \caption{Object Detection with \emph{Customized Class Set (e.g., question, reasoning text, open-vocabulary description)}}
    \label{fig:2a}
\end{subfigure}    
\hspace{0.1in}
\begin{subfigure}{0.48\textwidth}
     \centering
     \includegraphics[width=0.96\textwidth]{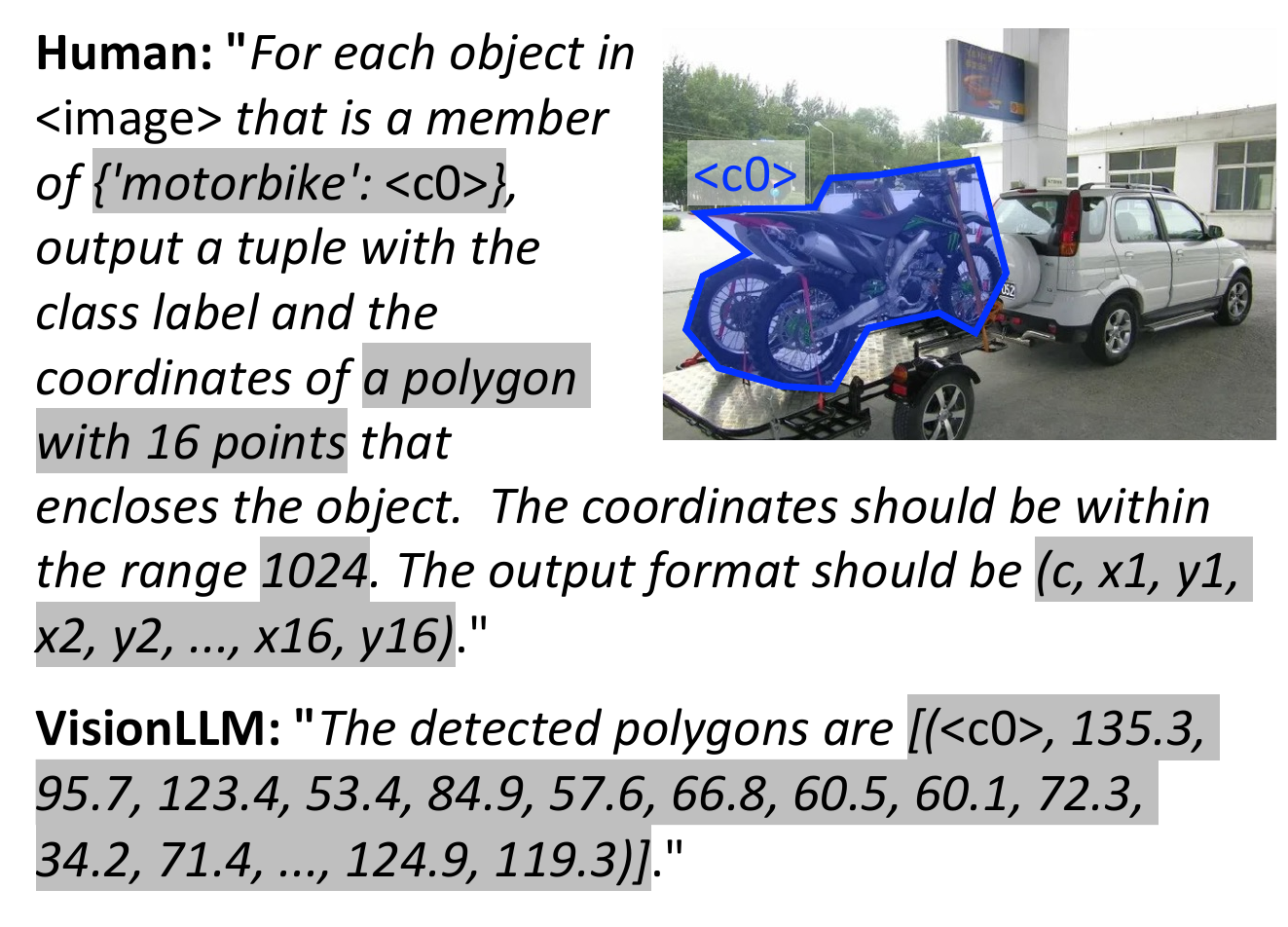}
     \caption{Object Location with \emph{Editable Output Format}}
     \label{fig:2b}
\end{subfigure}
\begin{subfigure}{0.48\textwidth}
     \centering
     \includegraphics[width=0.96\textwidth]{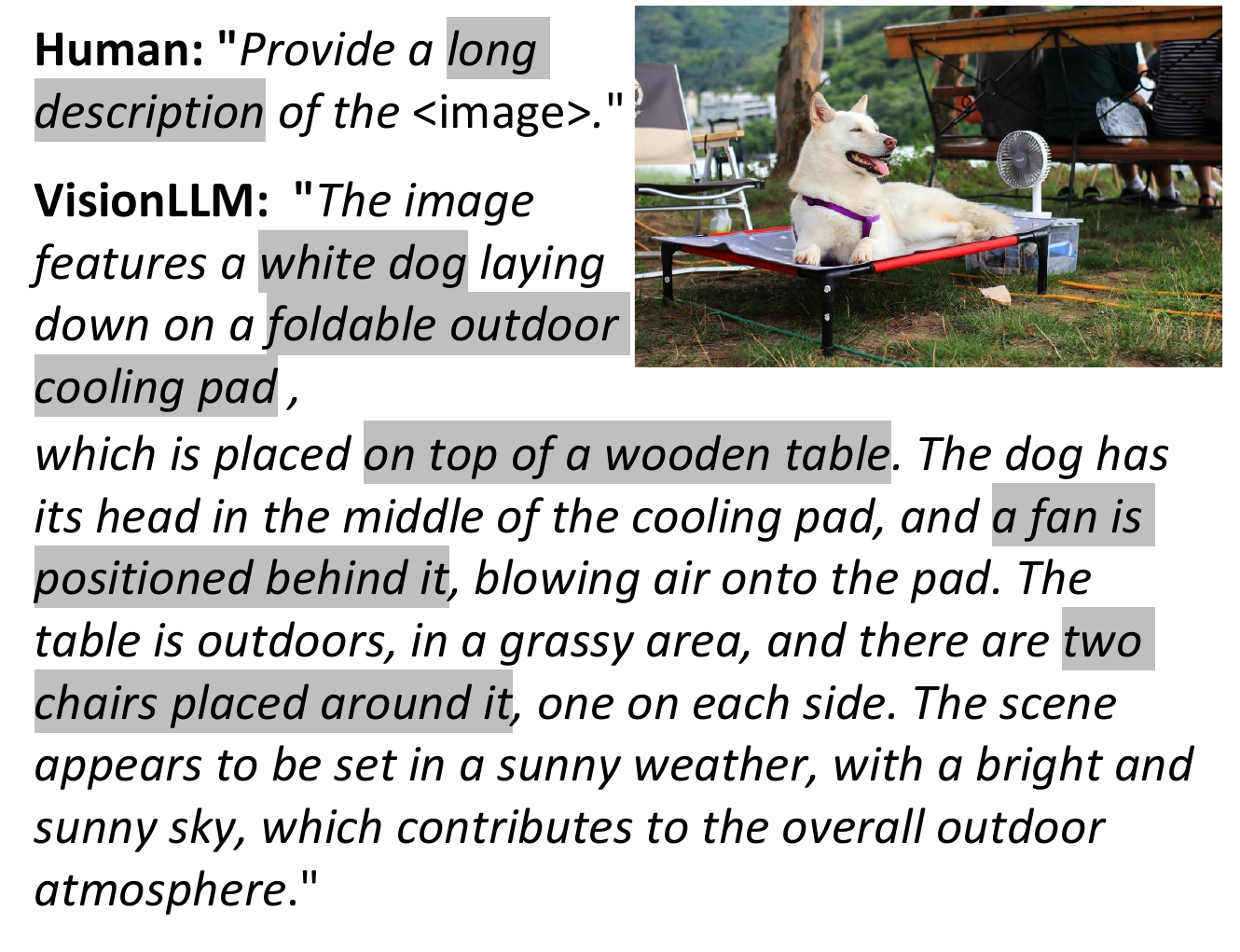}
     \caption{Image Description with \emph{Controllable Text Length}}
     \label{fig:2c}
\end{subfigure}
\hspace{0.1in}
\begin{subfigure}{0.48\textwidth}
     \centering
     \includegraphics[width=0.955\textwidth]{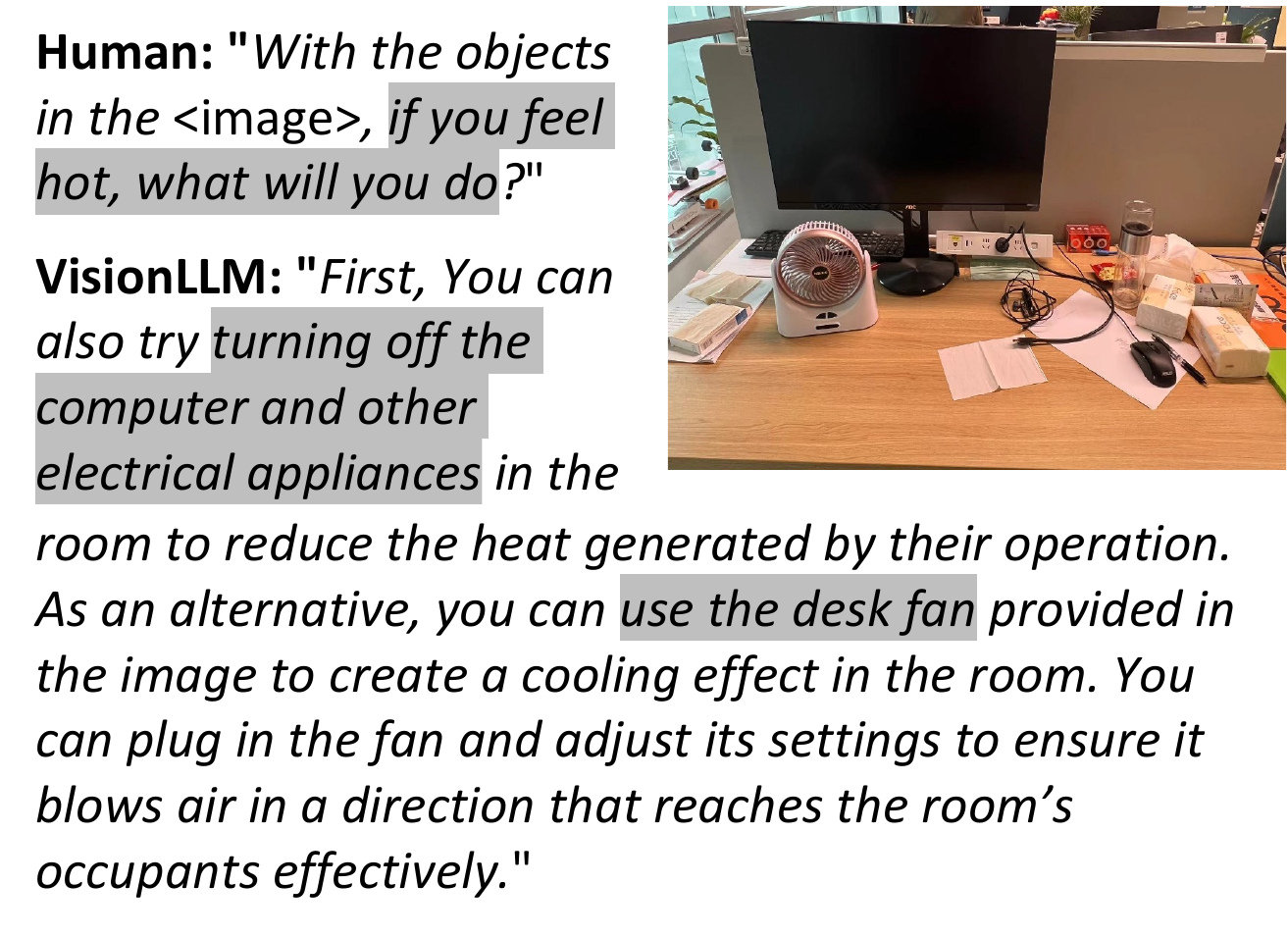}
     \caption{Visual Question Answer with \emph{Complex Reasoning}}
     \label{fig:2d}
\end{subfigure}
\caption{\textbf{Results and visualizations of our VisionLLM}. Guided by language instructions, our unified generalist framework showcases its effectiveness on diverse open-ended vision-centric tasks.
The text marked with a gray background indicates the customized instructions and the desired outputs.
}
\label{fig:res}
\end{figure}

(3) We construct a series of tasks with different granularities to verify the effectiveness of our models, ranging from easy to hard, and from pre-defined to flexible. Through these validations, we demonstrate the remarkable generality of our models, showcasing their ability to handle diverse scenarios, including random object categories, random output formats, and random task descriptions, as shown in Figure \ref{fig:res}.
The successful outcomes of these validations underscore the tremendous potential of our model in harnessing the capabilities of LLMs to control and guide vision-centric tasks.
In addition, with a generalist LLM-based framework, our model also yields promising results on various vision-centric tasks. Notably, our generalist model achieves an impressive mAP score of 60+\% on the COCO dataset, surpassing many detection-specific models~\cite{zhu2021deformable,carion2020end,he2017mask} and approaching the state-of-the-art record.

\section{Related Work}
\subsection{Large Language Model}

Large language models (LLMs) have gained significant attention in the field of natural language processing (NLP) and artificial general intelligence (AGI), due to their impressive capabilities in language generation, in-context learning, world knowledge, and reasoning.
The GPT family, including GPT-3 \cite{brown2020gpt3}, ChatGPT~\cite{openai2022chatgpt}, GPT-4~\cite{openai2023gpt4}, and InstructGPT \cite{ouyang2022instruct-tuning} are most representative works of LLMs. 
Other LLMs like OPT~\cite{zhang2022opt}, LLaMA~\cite{touvron2023llama}, MOSS~\cite{moss}, and GLM~\cite{zeng2022glm} have also made substantial contributions to the field.
These models achieve high performance and are open-sourced, serving as valuable resources for training large models and as foundations for further fine-tuning for specific purposes. 
For instance, Alpaca~\cite{taori2023alpaca} introduces a self-instruct framework that facilitates instruction tuning of the LLaMA model, reducing the reliance on human-written instruction data.
Recently, the emergence of these LLMs has also opened up API-based applications for solving vision-centric tasks. 
These applications have integrated visual APIs with language models to enable decision-making or planning based on visual information, such as Visual ChatGPT~\cite{wu2023visual}, MM-REACT~\cite{Yang2023MMREACTPC}, HuggingGPT~\cite{shen2023hugginggpt}, InternGPT~\cite{liu2023interngpt}, and VideoChat~\cite{li2023videochat}.
However, despite the convenience of using language-based instructions to define tasks and describe visual elements, these interactive systems~\cite{wu2023visual, Yang2023MMREACTPC, shen2023hugginggpt,liu2023interngpt,li2023videochat} still face limitations in capturing fine-grained visual details and understanding complex visual contexts, which hinder their ability to effectively connecting vision and language models. 
In summary, while LLMs have shown tremendous potential in various NLP applications, their applicability to vision-centric tasks has been limited by the challenges posed by modalities and task formats.

\subsection{Vision Generalist Model}
The pursuit of generalist models~\cite{zhu2022uni_p, unifiedio, yan2023universal}, which aim to handle a wide range of tasks using a shared architecture and parameters, has been a long-standing goal in the machine learning community.
Inspired by the success of sequence-to-sequence (seq2seq) models in the field of NLP \cite{radford2018improving}, recent advancements such as OFA \cite{ofa}, Flamingo~\cite{alayrac2022flamingo}, and GIT~\cite{wang2022git} propose modeling diverse tasks as sequence generation tasks.
Unified-IO~\cite{unifiedio}, Pix2Seq v2~\cite{pix2seqv2}, and UniTab~\cite{unitab}  extend this idea by using discrete coordinate tokens to encode and decode spatial information for more tasks.
Gato~\cite{gato} also incorporates reinforcement learning tasks into the seq2seq framework, while GPV~\cite{gpv1} develops a general-purpose vision system by combining a seq2seq module with a DETR-based visual encoder~\cite{carion2020end}.
However, these methods suffer from some limitations, such as slow inference speed and performance degradation due to the non-parallel auto-regressive decoding process. Uni-Perceivers~\cite{zhu2022uni_p,zhu2022uni,li2022uni} solve these issues by unifying different tasks using the maximum likelihood target for each input based on representation similarity, regardless of their modality, making it possible to support both generation and non-generation tasks in a unified framework. 
Nevertheless, these generalist models are still restricted by pre-defined tasks and cannot support flexible open-ended task customization based on language instructions like LLMs. 

\subsection{Instruction Tuning}
Language instructions are a powerful way to express various NLP tasks and examples for LLMs, as introduced by GPT-3~\cite{brown2020gpt3}.
Following this idea, subsequent works, such as InstructGPT~\cite{ouyang2022instruct-tuning}, FLAN~\cite{chung2022scaling, wei2021finetuned}, and OPT-IML~\cite{iyer2022opt-iml}, explore the instruction-tuning method~\cite{wang2022benchmarking, wang2022self_instruct} and demonstrate that this simple approach effectively enhances the zero-shot and few-shot capabilities of LLMs.
The language instruction paradigm has also been adopted by the computer vision community to define image-to-text tasks.
Flamingo~\cite{alayrac2022flamingo} is a milestone work that uses vision and language inputs as prompts and achieves remarkable few-shot results in various vision-language tasks, such as image captioning~\cite{chen2015coco-caption} and VQA~\cite{antol2015vqa}.
BLIP-2~\cite{li2023blip-2} further connects the visual encoder with LLMs through a querying transformer and a linear projection layer to build strong multimodal models.
MiniGPT-4~\cite{zhu2023minigpt} and LLaVA~\cite{liu2023llava} finetune the BLIP-2-style models on synthetic multimodal instruction-following data to unleash the potential of LLMs.
However, these models mainly focus on image-to-text tasks and fail to address visual perception, such as object detection, instance segmentation, pose estimation, etc.
To tackle image inpainting tasks, Bar \etal\cite{bar2022visual} introduces the first visual prompting framework that utilizes inpainting with discrete tokens on images. Painter~\cite{wang2022painter} and SegGPT~\cite{wang2023seggpt} employ masked image modeling on raw pixels for in-context learning with paired images. While these visual prompt models demonstrate good results in segmentation tasks, their applicability to numerous real-world vision tasks is challenging. Moreover, defining the visual prompts as image inpainting is inconsistent with the language instructions in LLMs, hard to leverage the reasoning, parsing ability, and world knowledge of LLMs.
In this work, we aim to align vision-centric tasks with language tasks, use language instructions to unifiedly and flexibly define all tasks, and solve them with a shared LLM-based task decoder.

\section{VisionLLM}
\subsection{Overall Architecture}

\begin{figure}[t]
    \centering
    \includegraphics[width=0.95\textwidth]{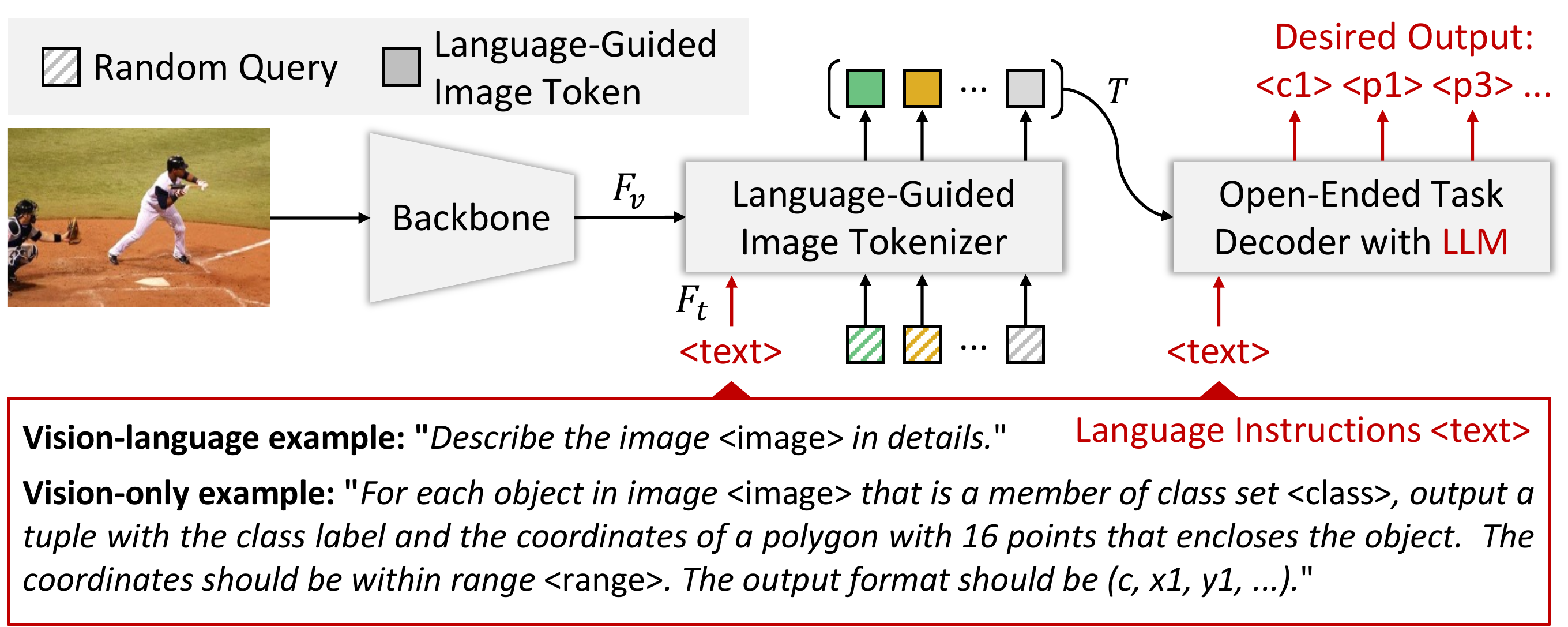}
    \caption{\textbf{Overall architecture of the proposed VisionLLM.} It consists of three parts: a unified language instruction designed to accommodate both vision and vision-language tasks, an image tokenizer that encodes visual information guided by language instructions, and an LLM-based open-ended task decoder that executes diverse tasks defined by language instructions.
    } 
    \label{fig:pipeline}
\end{figure}

This work targets to provide a unified generalist framework that can seamlessly integrate the strengths of large language models (LLMs) with the specific requirements of vision-centric tasks. 
As shown in Figure \ref{fig:pipeline}, the overall architecture of VisionLLM consists of three key designs: (1) a unified language instruction that provides a consistent interface for vision-centric task definition and customization; (2) a language-guided image tokenizer, which encodes visual information in alignment with the given language prompt, enabling the model to comprehend and parse the visual content effectively; and (3) an LLM-based open-task decoder, which utilizes the encoded visual information and language instructions to generate satisfactory predictions or outputs. The three designs work together to achieve a flexible and open-ended framework that can handle various vision-centric tasks at different levels of task customization through language instructions.

Different from previous interactive systems \cite{wu2023visual, Yang2023MMREACTPC, shen2023hugginggpt,liu2023interngpt,li2023videochat} that rely on APIs, our VisionLLM presents a more flexible and end-to-end pipeline.
Given language instructions that describe the current tasks and an input image, the model first uses a language-guided image tokenizer to encode the image tokens based on the given prompt. Then, the image tokens and 
language instructions are fed to an LLM-based open-ended task decoder. Finally, it evaluates the generated outputs against the task definition given by the unified language instructions, enabling the model to produce task-specific results. This seamless, end-to-end pipeline enables VisionLLM to effectively combine vision and language, achieving remarkable performance in open-ended and customizable vision-centric tasks.

\subsection{Unified Language Instruction}
\label{sec:language_instruction}
We first  introduce unified language instructions to describe vision-centric tasks. This design enables the unification of various vision-only and vision-language task descriptions and allows for flexible task customization.

\noindent \textbf{Vision-Language Tasks.} The instructions for vision-language tasks such as image captioning and visual question answering (VQA) are straightforward and similar to NLP tasks. 
Following previous methods~\cite{li2023blip-2,zhu2022uni_p,liu2023llava},
we describe the image captioning task like 
``\textit{The image is } $\texttt{<image>}$. \textit{Please generate a caption for the image: }'', and the VQA task like ``\textit{The image is } $\texttt{<image>}$. \textit{Please generate an answer for the image according to the question: }{$\texttt{<question>}$}''. Here, $\texttt{<image>}$ and $\texttt{<question>}$ are the placeholdersok of the image tokens and the question, respectively.

\noindent \textbf{Vision-Only Tasks.} 
Designing effective language instructions for vision tasks is a challenging endeavor due to the differences in modality and task format between vision and language.
Here, we describe vision tasks by providing a task description and specifying the desired output format via language instructions.

(1) The task description conveys the intended task to the language model. Following self-instruct~\cite{wang2022self_instruct},  we design a set of seed instructions with placeholders and employ 
LLMs
to generate a large number of related task descriptions and randomly select one of them during training.

(2) For conventional visual perception tasks like object detection and instance segmentation, we propose a unified output format represented as a tuple $(C, P)$, where $C$ denotes the class index in the category set $\texttt{<class>}$, 
and $P\!=\!\{x_i, y_i\}_{i=1}^N$ represents $N$ points that locate the object. 
To align with the format of word tokens, both the class index $C$ and the coordinates of points $x_i, y_i$ are transformed into discretized tokens. Specifically,  the class index is an integer starting from 0, and the continuous coordinates of the points are uniformly discretized into an integer within the range [-$\texttt{<range>}$, $\texttt{<range>}$].
For object detection and visual grounding tasks, the point number $N$ is equal to 2, representing the the top-left and bottom-right points of object's bounding box. In the case of instance segmentation, we employ multiple ($N\!>\!8$) points along the object boundary to represent an instance mask~\cite{xie2020polarmask}. Other perception tasks such as pose estimation (keypoint detection) can also be formulated as language instructions in this way.

An example of language instruction for the instance segmentation task is as follows:
``\textit{Segment all the objects of category set} \texttt{<class>} \textit{within the} \texttt{<range>} \textit{of the image and generate a list of the format (c, x1, y1, x2, y2, ..., x8, y8). Here, c represents the index of the class label starting from 0, and (x1, y1, x2, y2, ..., x8, y8) correspond to the offsets of boundary points of the object relative to the center point. The image is:} \texttt{<image>}''.

\subsection{Language-Guided Image Tokenizer}
\label{sec:image_tokenizer}
VisionLLM considers images as a kind of foreign language and converts them into token representations. 
Unlike previous works~\cite{dosovitskiy2021image,wang2022pvt,liu2021swin} that utilize fixed-size patch embeddings to represent images,
we introduce the language-guided image tokenizer to flexibly encode visual information that aligns with task-specific language prompts or instructions.

Specifically, give an image $\mathbf{X}\!\in\!\mathbb{R}^{H\!\times\!W\times\!3}$ with height $H$ and width $W$, we first feed it to the image backbones (\textit{e.g.}, ResNet~\cite{he2016deep}) and extract visual features $F_v$ of four different scales.
Additionally, we leverage a text encoder (\textit{e.g.}, BERT~\cite{devlin2018bert}) to extract the language features $F_l$ from given prompts. 
The language features are then injected into each scale of visual features through cross-attention~\cite{vaswani2017attention}, yielding multi-scale language-aware visual features, enabling the alignment of features across modalities.

Afterward, we propose to adopt a transformer-based network (\eg, Deformable DETR~\cite{zhu2021deformable}) with $M$ random-initialized queries $Q\!=\!\{q_{i}\}_{i=1}^{M}$ to capture the high-level information of images. We build the transformer-based network on top of the multi-scale language-aware visual features to extract $M$ image tokens $T\!=\!\{(e_i, l_i)\}_{i=1}^{M}$, each of which is represented by an embedding $e_i$ and a location $l_i$, denoting the semantic and positional information of the token. This design not only represents the images independent of input resolution but also extracts the visual representation that is informative with respect to the language prompts.

\subsection{LLM-based Open-Ended Task Decoder}
\label{sec:task_decoder}

We build our decoder on Alpaca~\cite{taori2023alpaca}, an LLM that is adapted from LLaMA~\cite{touvron2023llama}, to handle various vision-related tasks with language guidance. However, Alpaca has some inherent drawbacks for vision-centric tasks, such as (1) It only has a few digit tokens (\eg, 0$\sim$9) in its vocabulary, which restricts its ability to locate objects by numbers; (2) It uses multiple tokens to represent the category name, resulting in an inefficient scheme in object classification; and (3) It is a causal model that is inefficient for visual perception tasks.

To tackle these issues, we expand the vocabulary of LLM with additional tokens specially designed for vision-centric tasks.
First, we add a set of location tokens, denoted as \{$\texttt{<p-512>}$, ..., $\texttt{<p0>}$, ..., $\texttt{<p512>}$\}, 
where $\texttt{<p}\ \texttt{i>}$ represents the discretized offset of $i\!\in\![-512, 512]$ to the location $l_i$ of the image token, and the relative value to image height or width is equal to $i/512$.
These tokens successfully transform the object localization task from continuous variable prediction to more unified discrete bin classification.
Second, we introduce semantics-agnostic classification tokens \{$\texttt{<c0>}$, $\texttt{<c1>}$, ..., $\texttt{<c511>}$\} to replace category name tokens, which overcomes the inefficiency of using multiple tokens to represent categories.
The mapping between category names and the classification tokens is flexibly provided in the category set $\texttt{<class>}$ of language instructions, such as $\{\texttt{"person":<c0>}, \texttt{"car":<c1>}, \texttt{"black cat":<c2>,}...\}$. 
This design allows our model to select the appropriate category name from the provided category set, facilitating
efficient and accurate object classification.

\begin{wrapfigure}{r}{0.5\textwidth}
    \raggedleft
    \setlength{\fboxrule}{0pt}
    		\fbox{\includegraphics[width=0.5\textwidth]{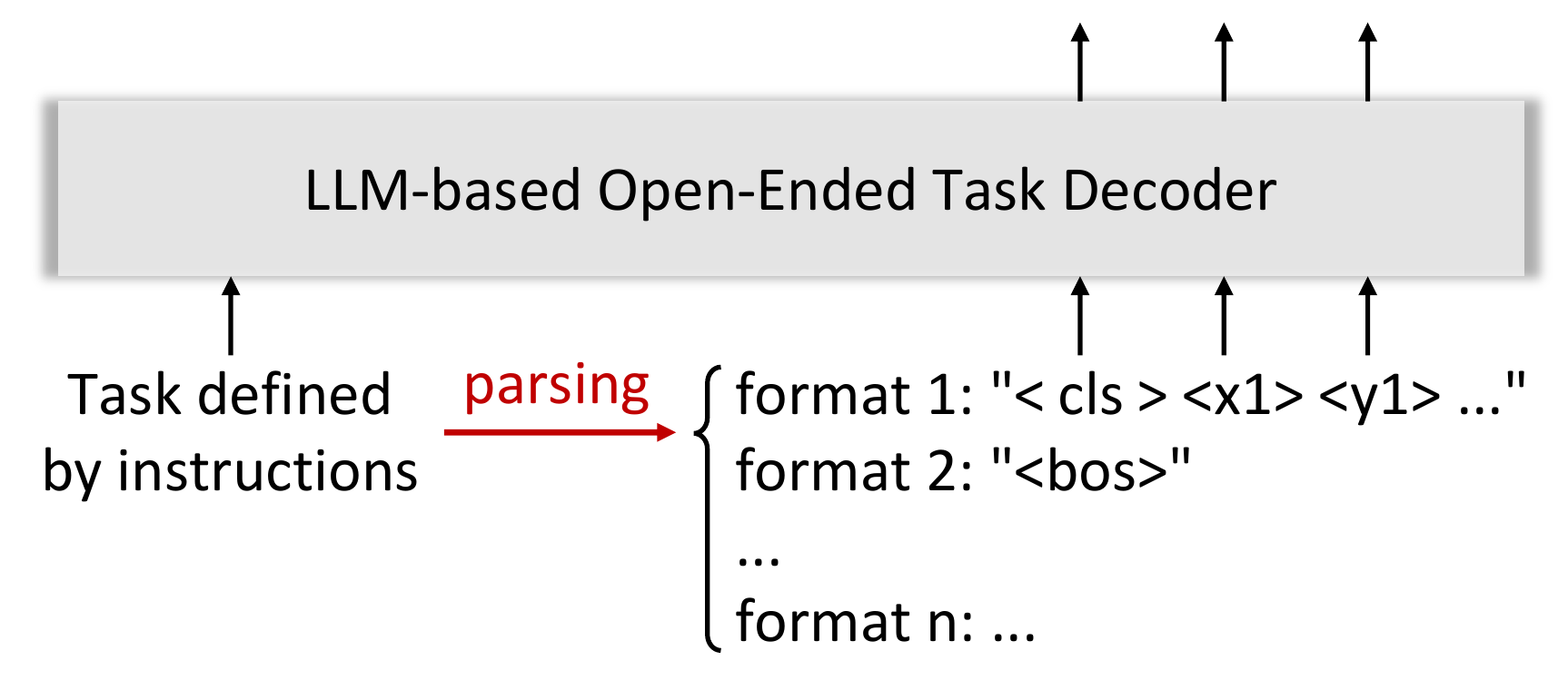}}
    \caption{
    Illustration of the ``output-format-as-query'' decoding process. ``\texttt{<cls>} \texttt{<x1>} \texttt{<y1>} ...'' denote the queries of the object's class index and boundary points, and
    ``\texttt{<bos>}'' denotes the beginning of string.
    }
    \label{fig:cloze}
\end{wrapfigure}
Moreover, to address the inefficiency caused by the causal framework, we introduce output-format-as-query decoding. We first use LLMs to parse the structural output format from the task instructions (\eg, ``\texttt{<cls> <x1> <y1> <x2> <y2>}'' for object detection, ``\texttt{<bos>}'' for image captioning), and then feed the tokens of structural output format as queries to the decoder to generate the desired output according to the queries. This simple method enables our model to not only avoid inefficient token-by-token decoding in visual perception tasks, but also keep a unified framework for vision-language tasks.

In this way, the output of object location and classification is formulated as a foreign language, thus unifying these vision-centric tasks into the format of token classification. Therefore, both vision-language and vision-only tasks can be supervised with the cross entropy loss like language tasks.
In addition, for efficient training, we adopt the Low-Rank Adaptation (LoRA) approach \cite{hu2021lora}, which allows us to train and fine-tune the models without excessive computational costs. 
It also acts as a bridge between the language and visual tokens, facilitating effective alignment between the two modalities, ensuring better task customization, and improving the convergence of the overall system.

\section{Experiment}

\subsection{Experimental Settings}

\textbf{Datasets.}
\modelname unifies the output formats of vision and language tasks as vocabulary generation, which enables models to be jointly trained on a wide range of tasks. In the experiments, we investigate the general modeling capacities of \modelname on five vision-centric tasks, including object detection, instance segmentation, visual grounding, image captioning, and visual question answering. For object detection and instance segmentation, COCO2017~\cite{lin2014microsoft} is used for training and evaluation. For visual grounding, we combine the annotations of RefCOCO~\cite{yu2016refcoco}, RefCOCO+~\cite{yu2016refcoco} and RefCOCOg~\cite{mao2016refcocog} for training, resulting in over 120k referred objects in total. And our models are evaluated on the validation set of RefCOCO. 
For image captioning and visual question answering, we adopt COCO Caption~\cite{chen2015coco-caption} and LLaVA-Instruct-150K~\cite{liu2023llava} as the training source. 
We evaluate the image captioning performance on the COCO Karpathy test split following common practice \cite{li2022uni, ofa, unitab}. 
We mainly use qualitative results (see Figure~\ref{fig:2d}) to demonstrate the VQA capability of our model, as LLaVA-Instruct-150K is not compatible with the standard VQA benchmark.
These tasks differ in their granularity, ranging from coarse-grained image level to fine-grained pixel level, enabling a comprehensive evaluation of the model's ability to adapt to different levels of customization through language instructions.

\textbf{Implementation Details.} 
We implement two variants of \modelname with two image backbones, \ie, ResNet~\cite{he2016deep} and InternImage-H~\cite{wang2022internimage}. 
For the language-guided image tokenizer, we adopt BERT-Base~\cite{Bertasius2021IsSA} as the text encoder and Deformable DETR (D-DETR)~\cite{zhu2021deformable} to capture high-level information. We set the number of queries $M$ to 100, and the number of encoder/decoder layers to 6 for D-DETR.
For the LLM, we employ Alpaca-7B~\cite{taori2023alpaca}, a LLaMA \cite{touvron2023llama} model fine-tuned with instructions, and equip it with LoRA~\cite{hu2021lora} for parameter-efficient fine-tuning.

The model is trained in two stages.
In the first stage, we initialize the model with the pre-trained weights of D-DETR, BERT, and Alpaca-7B, and train the visual backbone and the language-guided image tokenizer, while freezing most parameters of the LLM except a few LoRA parameters. 
To simplify the training complexity, in this stage, we mainly focus on object detection tasks with random object categories and task descriptions.
In the second stage, we freeze the visual backbone and introduce the unified supervision of multiple tasks.
Unless otherwise specified, the training runs for $50$ epochs on $4 \times 8$ NVIDIA A100 GPUs. AdamW \cite{adamw} is used as the optimizer, with one sample per GPU. We employ the cosine annealing schedule~\cite{loshchilov2016sgdr} as the learning policy, with an initial learning rate of $2 \times 10^{-4}$. In addition to the experiments in the main paper, more experimental settings and ablation studies are provided in the supplementary material due to space limitations.

\subsection{Task-Level Customization}
We first evaluate the task-level customization capability of \modelname. 
\modelname supports coarse-grained task customization, including visual perception tasks and visual-language tasks.
Table \ref{tab:benchmark} presents the evaluation results on four standard vision-centric tasks, including object detection, instance segmentation, visual grounding, and image captioning.  We compare our model with task-specific methods as well as recently-proposed vision generalist models. Note that, unless specifically mentioned, \emph{the results of our model come from a shared-parameter generalist model and switch different tasks by changing the language instructions only. Detailed instructions could be found in the supplementary material.}

\begin{table*}[t]
\centering 
\renewcommand\arraystretch{0.95} 
\setlength{\tabcolsep}{0.55mm}    
\footnotesize

\caption{\textbf{Results on standard vision-centric tasks.} 
`Intern-H'' denotes InternImage-H~\cite{wang2022internimage}.
``sep'' indicates that the model is separately trained on each task.}
\begin{tabular}{lcccccccccccccc}
      \shline
      \multirow{2}{*}{Method} & \multirow{2}{*}{Backbone} & 
       \multirow{2}{*}{\begin{tabular}[c]{@{}c@{}}Open-\\ Ended\end{tabular}} & 
      \multicolumn{4}{c}{Detection} &  \multicolumn{3}{c}{Instance Seg.} & \multicolumn{1}{c}{Grounding} & \multicolumn{2}{c}{Captioning} \\
      \cmidrule(lr){4-7} \cmidrule(lr){8-10} \cmidrule(lr){11-11} \cmidrule(lr){12-13}
      &  &  &  \multicolumn{2}{c}{AP} & \multicolumn{1}{c}{AP$\rm_{50}$}  & \multicolumn{1}{c}{AP$\rm_{75}$} & \multicolumn{1}{c}{AP} & \multicolumn{1}{c}{AP$\rm_{50}$}  & \multicolumn{1}{c}{AP$\rm_{75}$}& \multicolumn{1}{c}{P@0.5} & \multicolumn{1}{c}{BLEU-4} & \multicolumn{1}{c}{CIDEr}  \\
      \hline
      \multicolumn{13}{l}{\small{\textbf{\emph{Specialist Models}}}}  \\
      Faster R-CNN-FPN~\cite{ren2015faster} & ResNet-50 & - & \multicolumn{2}{c}{40.3} &61.0&44.0& - & - & - & - & - & - \\
      DETR-DC5~\cite{carion2020end} & ResNet-50 & - & \multicolumn{2}{c}{43.3} &63.1&45.9& - & - & - & - & - & - \\
      Deformable-DETR \cite{zhu2021deformable} & ResNet-50 & - & \multicolumn{2}{c}{45.7} &65.0&49.1& - & - & - & - & - & -\\
      Mask R-CNN~\cite{he2017mask} & ResNet-50 & - & \multicolumn{2}{c}{41.0} &61.7&44.9&37.1&58.4&40.1& - & - & -\\
      Polar Mask~\cite{xie2020polarmask}& ResNet-50 & - & \multicolumn{2}{c}{-} & - & - &30.5&52.0&31.1& - & - & - \\
      Pix2Seq \cite{chen2021pix2seq} & ResNet-50 & - & \multicolumn{2}{c}{43.2} & 61.0 & 46.1 & - & - & - & - & - & - \\
      UNITER \cite{chen2020uniter} & ResNet-101 & - &\multicolumn{2}{c}{-} & - & - & - & - & - &81.4&  - & -  \\
      VILLA~\cite{gan2020villa} & ResNet-101 & - &\multicolumn{2}{c}{-} & - & - & - & - & - &82.4&  - & -  \\
      MDETR~\cite{kamath2021mdetr} & ResNet-101 & - &\multicolumn{2}{c}{-} & - & - & - & - & - &86.8&  - & -  \\
      VL-T5 \cite{cho2021unifying} & T5-B  & - & \multicolumn{2}{c}{-} & - & - & - & - & - & - & - &116.5 \\
      \hline
      \multicolumn{13}{l}{\small{\textbf{\emph{Generalist Models}}}}  \\
      UniTab~\cite{yang2022unitab} & ResNet-101 & - & \multicolumn{2}{c}{-} & - & - & - & - & - &88.6& - &115.8 \\
      Uni-Perceiver \cite{zhu2022uni_p} & ViT-B & - & \multicolumn{2}{c}{-} & - & - & - & - & - & - &32.0& - \\
      Uni-Perceiver-MoE \cite{zhu2022uni} & ViT-B & - & \multicolumn{2}{c}{-} & - & - & - & - & - & - &33.2& - \\
      Uni-Perceiver-V2 \cite{li2022uni} & ViT-B & - & \multicolumn{2}{c}{58.6} & - & - &50.6& - & - & - &35.4&116.9 \\    
      Pix2Seq v2 \cite{pix2seqv2} & ViT-B & - & \multicolumn{2}{c}{46.5} & - & - &38.2& - & - & - &34.9& - \\ 
      
      \hline
      \rowcolor{lightgray!28}\modelname-R50$_{\rm sep}$& ResNet-50 & - & \multicolumn{2}{c}{44.8} & 64.1 & 48.5 & 25.2 & 50.6 & 22.4 & 84.4& 30.8 & 112.4  \\
      \rowcolor{lightgray!28}\modelname-R50 & ResNet-50 & \yes & \multicolumn{2}{c}{44.6} & 64.0 & 48.1 & 25.1& 50.0 & 22.4 & 80.6 & 31.0 & 112.5  \\
      \rowcolor{lightgray!28}\modelname-H & Intern-H & \yes &\multicolumn{2}{c}{60.2} & 79.3 & 65.8 & 30.6 & 61.2 & 27.6 & 86.7 & 32.1 & 114.2 \\
      \shline
\end{tabular}
\label{tab:benchmark} 
\end{table*}

\textbf{Object Detection.}
Object detection is a fundamental computer vision task that involves identifying and localizing objects of interest within an image. 
Our method achieves comparable or higher results to others, $44.6$ mAP, with a ResNet-50~\cite{he2016deep} backbone. 
With the same backbone \ie~ResNet-50, our method outperforms Pix2Seq~\cite{chen2021pix2seq} by $1.4$ mAP, which also discretizes the output coordinates to integers.
Furthermore, benefiting from the output-format-as-query framework (see Sec. \ref{sec:task_decoder}), we can decode multiple predictions in parallel during inference, making our approach more efficient. 
Using InternImage-H~\cite{wang2022internimage} as the visual backbone, we obtained 60.2\% mAP, which is close to the current state-of-the-art detection-specific model~\cite{wang2022internimage}, demonstrating the scalability of our generalist model.

\textbf{Visual Grounding.} 
Visual grounding associates textual descriptions with corresponding regions or objects within an image. Training visual grounding and object detection can potentially conflict with each other, 
as object detection aims to detect all the objects, while visual grounding should only localize the referred object and suppress other objects.
Benefiting from our unified task instructions and the strong instruction comprehension capabilities of LLMs, our model performs both tasks effectively and achieves a result of $80.6$ P@0.5 for visual grounding. 
With InternImage-H as the backbone, we achieve $86.7$ P@0.5 on the validation set of RefCOCO.

\textbf{Instance Segmentation.}
Instance segmentation involves identifying and segmenting individual objects within an image.
We employ a flexible number of points (\ie, 8$\sim$24) along the object boundary to represent an instance mask. 
Compared to mainstream models specific to instance segmentation, our model has a comparable mask AP$_{50}$ (61.2\% with InternImage-H~\cite{wang2022internimage}) but relatively low mask AP$_{75}$.
This gap could potentially arise from factors as follows: 
(1) We discretize the output coordinates to integers for unifying tasks, which introduces information loss; (2) Due to the memory and computational constraint, the number of points in our model is limited, which also results in a performance drop; and (3) Point-based methods typically yield lower results compared to direct mask prediction methods, such as Mask R-CNN~\cite{he2017mask}.

\textbf{Image Captioning.}
We also evaluate our model in a representative vision-language task, \ie~image captioning task, and report the BLEU-4~\cite{papineni2002bleu} and CIDEr~\cite{vedantam2015cider} metrics. Note that we do not adopt the beam search~\cite{anderson2018bottom} or CIDEr optimization~\cite{rennie2017self}. We can observe that \modelname achieves competitive performance to previous methods. 
With ResNet-50, we obtain a BLEU-4 score of $31.0$ and a CIDEr score of $112.5$. 
When using InternImage-H as the backbone, our model achieves a comparable BLEU-4 score of $32.1$ and a CIDEr score of $114.2$. 
These results demonstrate the effectiveness of \modelname in generating descriptive and contextually relevant captions for images. 

\subsection{Object-Level \& Output Format Customization}

Our VisionLLM not only allows for customizing the task description, but also for adjusting the target object and the output format using language instructions. Here, we evaluate our model’s fine-grained customization ability on COCO. In particular, to customize the target object, we modify the \texttt{<class>} in language instructions to change the model’s recognition target from $10$ classes to $80$ classes. Likewise, to customize the output format, we modify the number of points in language instructions to change the task output format. Table \ref{tab:customization} shows that our method can perform well for both object-level and output format changes.

\begin{table}[t]
\renewcommand\arraystretch{0.95} 
\centering
\small
\caption{\textbf{Experiments of object-level and output format customization.}
We conduct these experiments based on \modelname-R50, and report the performance of box AP  and mask AP on COCO minival for (a) and (b), respectively. ``\#Classes'' and ``\#Points'' indicate the number of classes and boundary points, respectively. ``*'' indicates that we report the mean AP of the given classes, \eg, 10 classes.
}
\label{tab:customization}
{
    \begin{minipage}{0.48\linewidth}{
    \subcaption{Object-level customization.}
    \begin{center}
    \setlength{\tabcolsep}{1mm}
    \begin{tabular}{ccccccc}
    \shline
     \#Classes  & AP  & AP$_{\rm 50}$& AP$_{\rm 75}$&  AP$_{\rm S}$& AP$_{\rm M}$& AP$_{\rm L}$  \\
    \hline
     10$^*$   & 48.9 & 72.6 & 51.2 & 31.7 &  47.5 & 67.3 \\
     20$^*$   & 52.7 & 73.6 & 56.8 & 31.8 & 53.2 & 70.5 \\
     40$^*$   & 49.3 & 70.7 & 53.2 & 33.1 & 53.6 & 63.8 \\
     80$^*$   & 44.6 & 64.0 & 48.1 & 26.7 & 47.9 & 60.5 \\
    \shline
    \end{tabular}
    \label{tab:abla1}
    \end{center}}
    \end{minipage}
}
{
    \centering
    \begin{minipage}{0.48\linewidth}{
    \subcaption{Output format customization.}
    \setlength{\tabcolsep}{1.0mm}
        \begin{center}
            \begin{tabular}{ccccccc}
            \shline
            \#Points & AP  & AP$_{\rm 50}$& AP$_{\rm 75}$&  AP$_{\rm S}$& AP$_{\rm M}$& AP$_{\rm L}$  \\
            \hline
            8     & 18.5 & 45.7 & 11.6 & 9.9 & 19.7 & 28.7 \\
            14    & 22.9 & 48.3 & 19.4 & 11.0 & 25.1 & 36.0 \\
            16    & 24.2 & 49.9 & 20.9 & 11.5 & 26.3 & 36.8 \\
            24    & 25.1 & 50.0 & 22.4 & 12.5 & 27.4 & 38.2 \\
            \shline
            \end{tabular}
            \label{tab:abla2}
        \end{center}}
    \end{minipage}
}
\end{table}

\subsection{Ablation Study}

In this section, we analyze the effect of key components and hyper-parameters on \modelname. Unless otherwise specified, we use ResNet-50~\cite{he2016deep} backbone and perform the ablation experiments for object detection tasks with random classes and task descriptions on COCO2017~\cite{lin2014microsoft}.

\begin{table}[t]
\centering
\footnotesize
\caption{\textbf{Ablation studies on language-guided image tokenizer and hyper-parameters.} 
}
\vspace{-5pt}
\label{tab:ablations}
{
    \centering
    \begin{minipage}{0.35\linewidth}{
    \subcaption{Effect of text encoder in the language-guided image tokenizer.}
    \setlength{\tabcolsep}{3pt}
    \renewcommand{\arraystretch}{0.95}
        \begin{center}
            \begin{tabular}{cccc}
                \shline
                w/ BERT & Freeze & COCO & RefCOCO \\
                \hline
                -   & -     & 44.7 & 48.1 \\
                \yes & -    & 44.8 & 84.1 \\
                \yes & \yes & 1.3  & 34.3 \\        
                \shline
        \end{tabular}
        \label{tab:ab_bert}
        \end{center}
        }
    \end{minipage}
}
\hspace{0.2in}
{
    \begin{minipage}{0.3\linewidth}{
    \begin{center}
    \subcaption{Effect of image tokenization method.}
    \setlength{\tabcolsep}{5pt}
    \renewcommand{\arraystretch}{0.95}
    \begin{tabular}{lc}
    \shline
     Tokenization    & AP \\
    \hline
     Average Pooling & 23.1 \\
     Ours            & 44.8 \\
    \shline
    \end{tabular}
    \label{tab:ab_tokenization}
    \end{center}}
    \end{minipage}
}
\hspace{0.2in}
{
    \begin{minipage}{0.2\linewidth}{
    \begin{center}
    \subcaption{Effect of the number of bins (\#Bins).}
    \setlength{\tabcolsep}{6pt}
    \renewcommand{\arraystretch}{0.95}
      \begin{tabular}{lc}
        \shline
         \#Bins  & AP   \\
      \hline
         257  & 34.9 \\
         513  & 40.8 \\
         1025 & 44.8 \\
         2049 & 44.8 \\
      \shline
    \end{tabular}
    \label{tab:ab_bin}
    \end{center}
    }
    \end{minipage}
}
\vspace{-10pt}
\end{table}

\textbf{Single Task \emph{vs.} Multiple Tasks.}
We perform an ablation study to assess the impact of multi-task learning with language instructions on \modelname. As shown in Table \ref{tab:benchmark}, the single-task trained model \modelname-R50$_{\rm sep}$ is slightly better than the jointly trained model \modelname-R50 except image captioning. This is due to the multitasking conflicts that also affect previous generalist models~\cite{zhu2022uni_p,zhu2022uni}, and it reflects a trade-off between accuracy and generalization.

\textbf{Text Encoder in Language-Guided Image Tokenizer.} 
We examine the role of text encoder (\ie, BERT) in our language-guided image tokenizer in Table \ref{tab:ab_bert}, where we report the results for object detection and visual grounding. The first two rows show that BERT is not essential for object detection but it is crucial for visual grounding. We also investigate the effect of freezing the text encoder during training. The last row indicates that freezing BERT hinders the alignment of vision and language modalities and thus degrades the performance for both tasks.

\textbf{Image Tokenization Method.} As a comparison to our query-based tokenization, we employ average pooling on the feature maps from the D-DETR encoder to obtain $M$ patch embeddings, which serve as token representations for the image. 
Results in Table \ref{tab:ab_tokenization} indicate a clear advantage of our method. 
This is due to its ability to capture information from objects of various sizes in a more flexible way.

\textbf{Number of Localization Tokens.} 
We vary the number of localization tokens from $257$ (\ie, -128$\sim$128) to $2049$ (\ie, -1024$\sim$1024), to investigate its impact on visual perception performance.
As presented in Table \ref{tab:ab_bin}, the model consistently exhibits improvement as the number of localization tokens increases until it reaches a saturation point. Remarkably, a substantial performance boost is observed when the number is raised from $257$ to $1025$ ($+9.9$ AP). 
These results indicate that a higher number of localization tokens enables the models to achieve finer localization abilities, thereby improving localization accuracy.

\section{Conclusion}
In this paper, we have presented VisionLLM, a novel framework that leverages the power of large language models (LLMs) to address vision-centric tasks in an open-ended and customizable manner. We have designed unified language instruction that matches the format of language models and covers various vision-centric tasks including visual perception. We have also developed a language-guided image tokenizer and an LLM-based task decoder that can handle open-ended tasks according to the given language instructions.
We have verified the effectiveness of our models on a series of tasks with different granularities, demonstrating their remarkable generality and flexibility.

\textbf{Broader Impact.} We envision that this work will promote the fusion of visual and language tasks. In addition, since our work is built on open-source pre-trained vision foundation models and large language models, requiring low training resources, thus reducing the carbon footprint. 
We do not foresee obvious undesirable ethical/social impacts at this moment.

\clearpage

\appendix

\Large{\textbf{Appendix}}
\normalsize

\section{Example Instructions}
As described in Sec.~\ref{sec:language_instruction} of the main paper, we follow self-instruct~\cite{wang2022self_instruct} to design a set of seed instructions with placeholders and employ LLMs to create diverse related task descriptions for coarse-grained task-level customization.
Here, we show some examples of instructions for task-level customization, including object detection, instance segmentation, visual grounding, image captioning, and visual question answering (VQA).
\emph{Following various instructions, our model can elegantly switch among different vision-centric tasks and accomplish them in a unified manner like LLMs.}  

\subsection{Object Detection}

{\textbf{Example 1.}} ``\textit{Please examine the image and identify all objects in the category set} \texttt{<class>}. \textit{For each object, specify its location within the range} \texttt{<range>} \textit{by determining the top-left and bottom-right corners of its bounding box. To indicate the object's class and location, provide the output in the format (c, x1, y1, x2, y2), where `c' represents the class index starting from 0, and (x1, y1, x2, y2) correspond to the offsets of the bounding box corners relative to the center point. The image is:} \texttt{<image>}''

{\textbf{Example 2.}} ``\textit{Identify all the objects in the image that belong to the category set} \texttt{<class>} \textit{and predict a bounding box around each one. The output should be a list in the format (c, x1, y1, x2, y2), where c represents the index of the class label starting from 0, and x1, y1, x2, y2 are the offsets of the top-left and bottom-right corners of the box relative to the center point. The coordinates should be within } \texttt{<range>}. \textit{The image is:} \texttt{<image>}''

{\textbf{Example 3.}} ``\textit{For each object in the image that is a member of the category set} \texttt{<class>}, \textit{output a tuple with the index of class label starting from 0 and the offsets of corners relative to the center point that encloses the object. The offsets should be in the order of top-left and bottom-right corners of the rectangle and should be within} \texttt{<range>}. \textit{The output format should be (c, x1, y1, x2, y2). The image is:} \texttt{<image>}''

\subsection{Instance Segmentation}
\label{sec:ins_seg_prompt}

{\textbf{Example 1.}} ``\textit{Segment the objects from the image with class labels from} \texttt{<class>} \textit{and output their coordinates within range} \texttt{<range>}. \textit{The coordinates should be given as the boundary points relative to the center point, and the output format should be (c, x1, y1, x2, y2, ..., x20, y20), where c is the index of the class label that starts from 0. The image is:} \texttt{<image>}''

{\textbf{Example 2.}} ``\textit{Segment all the objects from the category set} \texttt{<class>} \textit{in the provided image and output a tuple (c, x1, y1, x2, y2, ..., x14, y14) for each, where c is the index of the class label in the category set that starts from 0, and (x1, y1, x2, y2, ..., x14, y14) correspond to the offsets of boundary points on the instance mask relative to the center point which should be within} \texttt{<range>}. \textit{The image is:} \texttt{<image>}''

{\textbf{Example 3.}} ``\textit{In the provided image, please segment all the objects in category set} \texttt{<class>} \textit{within the range} \texttt{<range>} \textit{by providing their coordinates in the (c, x1, y1, x2, y2, ..., x24, y24) format, where `c' denotes the index of the class label starting from 0, and (x1, y1, x2, y2, ..., x24, y24) stand for the offsets of boundary points relative to the center point. The image is:} \texttt{<image>}''

\subsection{Visual Grounding}

{\textbf{Example 1.}} ``\textit{Please find the object in the category set} \texttt{\{<expression>:<cls0>\}} \textit{within the range} \texttt{<range>}. \textit{Please provide the output in the format (c, x1, y1, x2, y2), where c is the class index starting from 0, and (x1, y1, x2, y2) are the offsets of the top-left and bottom-right corners of the bounding box relative to the center point. The image is:} \texttt{<image>}''

{\textbf{Example 2.}} ``\textit{Given the input image, category set} \texttt{\{<expression>:<cls0>\}}, \textit{and the range} \texttt{<range>}, \textit{please locate the object in the image and output the corresponding coordinates in the tuple (c, x1, y1, x2, y2), where c is the index of the class label starting from 0, and (x1, y1, x2, y2) are the offsets of the top-left and bottom-right corners of the rectangle relative to the center point. The image is:} \texttt{<image>}''

{\textbf{Example 3.}} ``\textit{For each object in the image that belongs to the} \texttt{\{<expression>:<cls0>\}} \textit{category set, please provide the class label (starting from 0) and the offsets from the center of a bounding box that encloses the object. The corner offsets should be in the order of top-left and bottom-right, and within the range} \texttt{<range>}. \textit{The output should be in the format (c, x1, y1, x2, y2). The image is:} \texttt{<image>}''

\subsection{Image Captioning}

\textbf{Example 1.} ``\textit{The image is} \texttt{<image>}. \textit{Write a caption:} ''

\textbf{Example 2.} ``\textit{The image is} \texttt{<image>}. \textit{Please describe this image:} ''

\textbf{Example 3.} ``\textit{With the objects
in the} \texttt{<image>},  \textit{please generate a caption for the image:} ''

\subsection{Visual Question Answering}

\textbf{Example 1.} ``\textit{The image is} \texttt{<image>}. \textit{Please generate an answer according to the question: }{\texttt{<question>}}. ''

\textbf{Example 2.} ``\textit{The image is} \texttt{<image>}. \textit{Please answer the question} {\texttt{<question>}} \textit{according to the image}. ''

\textbf{Example 3.} ``\textit{With the objects
in the} \texttt{<image>}, {\texttt{<question>}}. ''

\section{Loss Function}
VisionLLM consists of two model components: language-guided image tokenizer and LLM-based open-task decoder. So the total loss $\mathcal{L}$ of our model can be written as:
\begin{equation}
    \mathcal{L} = \mathcal{L}_{\rm tok} + \mathcal{L}_{\rm dec},
\end{equation}
where $\mathcal{L}_{\rm tok}$ and $\mathcal{L}_{\rm dec}$ denote the loss of language-guided image tokenizer and LLM-based open-task decoder, respectively.
We introduce the two loss functions as follows:

\textbf{Language-Guided Image Tokenizer.} Different from the Q-Former~\cite{li2023blip-2}, we use a supervision method similar to that of Deformable DETR \cite{zhu2021deformable}, but with a different loss $\mathcal{L}_{\rm tok}$: category-agnostic classification (focal loss~\cite{lin2017focal}) and center point regression ($L_1$ loss). As explained in Sec.~\ref{sec:image_tokenizer}, our image tokenizer extracts $M$ image tokens $T\!=\!\{(e_i, l_i)\}_{i=1}^{M}$, each of which is represented by an embedding $e_i$ and a location $l_i$ (\ie, absolute coordinates of the center point).

\textbf{LLM-Based Open-Ended Task Decoder.} 
We handle two cases in decoding processing differently. (1) For regular word prediction, we train with standard next-token supervision~\cite{touvron2023llama,radford2018gpt1,brown2020gpt3,radford2019gpt2}; (2) For unordered set prediction (\eg, bounding boxes), we first output a sequence of tokens according to the output format (see the output-format-as-query paradigm in Sec.~\ref{sec:task_decoder}), then use bipartite matching to align the LLM-predicted outputs with the ground truths. Despite the differences, we use cross-entropy to compute the loss $\mathcal{L}_{\rm dec}$ in a unified way for both cases.

\begin{figure}[t!]
    \centering
    \includegraphics[width=0.85\textwidth]{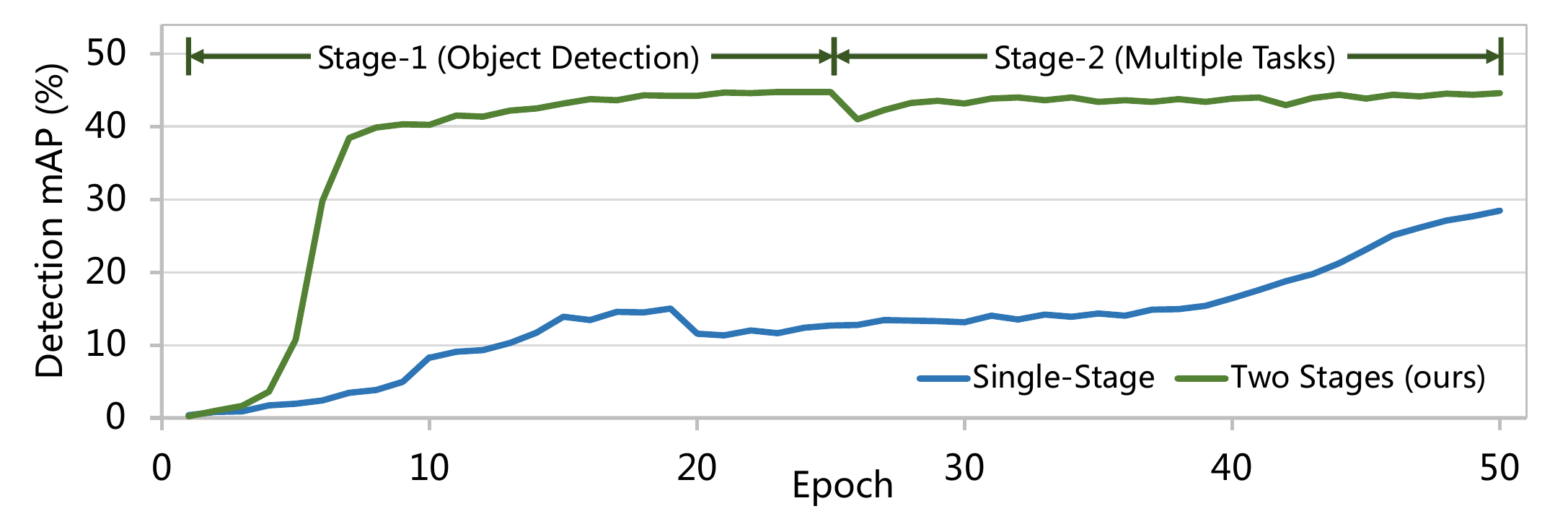}
    \caption{\textbf{Comparison of two training schedules for VisionLLM.} 
    We found that a two-stage training from easy to hard converges faster than a single-stage training.
    } 
    \label{fig:training_schedule}
\end{figure}

\section{Training Schedule}
\label{sec:training_schedule}

As shown in Figure \ref{fig:training_schedule}, to speed up the convergence of VisionLLM, we split the training schedule of VisionLLM into two stages:

\textbf{Stage 1.}
In this stage, we initialize the language-guided image tokenizer by loading the pre-trained weights of Deformable DETR \cite{zhu2021deformable} and BERT~\cite{devlin2018bert}. 
Additionally, Alpaca  \cite{taori2023alpaca} is employed as the LLM-based open-ended task decoder. 
To align visual tokens with text tokens, we make the language-guided image tokenizer trainable while freezing most parameters of the pre-trained Alpaca, with only a few LoRA \cite{hu2021lora} parameters left tunable. 
We only focus on object detection in this stage to simplify the training difficulty, with random task descriptions and object categories.

\textbf{Stage 2.} 
The second stage builds upon the model weights obtained from the first stage.
For efficiency, we freeze the visual backbone (\eg, ResNet \cite{he2016deep}) in the language-guided image tokenizer.
Notably, this stage introduces the unified supervision of multiple tasks, including object detection, instance segmentation, visual grounding, image captioning, and VQA, facilitating the model to leverage the power of LLMs to understand and manipulate visual information holistically.

\begin{table}[t]
\centering
\footnotesize
\caption{\textbf{More ablation studies for VisionLLM.}
}
\label{tab:ablations_appendix}
{
    \centering
    \begin{minipage}{0.35\linewidth}{
    \subcaption{Effect of randomness.}
    \setlength{\tabcolsep}{3pt}
    \renewcommand{\arraystretch}{0.95}
        \begin{center}
            \begin{tabular}{lcc}
                \shline
                Randomness         & AP      \\
                \hline
                None                  & 45.2    \\
                + Random Task Description & 45.1  \\  
                ++ Random Object Category & 44.8    \\
                +++ Random Output Format  & 44.6    \\  
                ~~~~~~~~(Multi-task Joint Training) \\
                \shline
        \end{tabular}
        \label{tab:ab_random}
        \end{center}
        }
    \end{minipage}
}
\hspace{0.1in}
{
    \begin{minipage}{0.3\linewidth}{
    \begin{center}
    \subcaption{Effect of LoRA \cite{hu2021lora}.}
    \setlength{\tabcolsep}{6pt}
    \renewcommand{\arraystretch}{0.95}
      \begin{tabular}{ccc}
        \shline
         LoRA  & Randomness & AP   \\
      \hline
         \no  & \no & 45.2 \\
         \no  & \yes & 1.2 \\
         \yes  & \yes & 44.8 \\
    \shline
          &  \\
          &  \\
    \end{tabular}
    \label{tab:effect_lora}
    \end{center}
    }
    \end{minipage}
}
\hspace{0.1in}
{
    \begin{minipage}{0.23\linewidth}{
    \begin{center}
    \subcaption{Effect of the number of image tokens.}
    \setlength{\tabcolsep}{5pt}
    \renewcommand{\arraystretch}{0.95}
    \begin{tabular}{cc}
    \shline
     \#Tokens    & AP \\
    \hline
     50     &  44.5 \\
     100    &  44.8 \\
     200    &  45.1 \\
     300    &  45.2 \\
    \shline
    \end{tabular}
    \label{tab:num_img_token}
    \end{center}}
    \end{minipage}
}
\hspace{0.1in}
{
    \begin{minipage}{0.23\linewidth}{
    \begin{center}
    \subcaption{Effect of Seq2Seq.}
    \setlength{\tabcolsep}{5pt}
    \renewcommand{\arraystretch}{0.95}
    \begin{tabular}{cc}
    \shline
     Seq2Seq    & AP \\
    \hline
     \yes     &  - \\
     \no    &  44.8 \\
    \shline
    \end{tabular}
    \label{tab:seq2seq}
    \end{center}}
    \end{minipage}
}
\hspace{0.1in}
{
    \begin{minipage}{0.4\linewidth}{
    \begin{center}
    \subcaption{Large vocabulary object detection.}
    \setlength{\tabcolsep}{5pt}
    \renewcommand{\arraystretch}{0.95}
    \begin{tabular}{ccc}
    \shline
     Dataset   & \#Classes &   AP   \\
    \hline
     COCO      & 80        &  44.8  \\
     LVIS      & 1203      &  18.9  \\
    \shline
    \end{tabular}
    \label{tab:lvis}
    \end{center}}
    \end{minipage}
}
\end{table}

\section{More Ablation Studies}

In this section, we provide more ablation studies and analysis of \modelname. Unless otherwise specified, we use ResNet-50~\cite{he2016deep} backbone and perform the ablation experiments for object detection tasks with random task descriptions and object categories on COCO 2017~\cite{lin2014microsoft}.

\textbf{Randomness.}
In Table \ref{tab:ab_random}, we examine the effect of introducing randomness during training for VisionLLM, including randomness in task descriptions, object categories, and output formats (\ie, multi-task joint training). Initially, without any randomness, the model achieves a box AP of $45.2$.
However, as randomness is gradually applied, interesting phenomena emerge: 
while there is a slight decrease ($45.2\rightarrow44.6$) in the AP of standard detection with the introduction of randomness, the overall benefits of enhanced task customization and open-ended capabilities outweigh this minor trade-off.
Overall, introducing randomness during training in VisionLLM positively impacts its capacity for open-ended tasks and customization.

\textbf{Low-Rank Adaptation (LoRA).}
As shown in Table \ref{tab:effect_lora}, when randomness is not applied, the model achieves $45.2$ box AP without using LoRA \cite{hu2021lora}. 
However, when randomness is employed, it is observed that the model fails to converge without using LoRA. 
Conversely, when LoRA and randomness are used together, the model is able to converge.
This indicates that LoRA plays a crucial role as a bridge between the language and visual tokens, enabling effective alignment between the two modalities and improving the convergence of the overall system.

\textbf{Number of Image Tokens.}
We vary the number of image tokens from $50$ to $300$ to investigate their impact on the performance. Results are presented in Table~\ref{tab:num_img_token}. As the number of image tokens increases, the performance continues to improve. This makes sense because a larger number of image tokens provides a more detailed description of the image content. Considering computational complexity, we adopted $100$ image tokens in our experiments.

\begin{figure}[t!]
\centering
\begin{tikzpicture}[font=\footnotesize]
\pgfplotsset{set layers, compat=1.8, 
every axis/.append style={
font=\scriptsize,
}
}
    \begin{groupplot}[
        group style={
            group size=1 by 2,
            vertical sep=0pt
        },
        axis lines = left,
        every outer y axis line/.style={draw=gray!40},
        every outer x axis line/.style={draw=gray!40},
        yticklabels={},
    	footnotesize,
        scale only axis,
    	x post scale=1.5,
        xticklabels={$1$, $2$, $3$, $4$, $5$, $6$, $7$, $8$},
    	xmin=1,
    	xmax=8,
    ylabel={AP},
    y label style={at={(0.08,0.16)},rotate=90,},
    xlabel={Prompt},
    nodes near coords, 
	nodes near coords align={vertical},
    	scaled ticks=false,
     ymajorgrids=true,
    ]
        \nextgroupplot[
          y post scale=0.4,
          xtick=data,
          ymin=40,
          ymax=50,
          ytick style={draw=none},
          enlargelimits=false,
    	  enlarge x limits=0.035,
    	  extra y tick style={grid=major, draw=gray!40},
          grid style={line width=.1pt, draw=gray!20},
          ylabel style = {rotate=270, font=\footnotesize, anchor=north, yshift=35pt, xshift=13.5pt}
        ]
        \addplot[draw=mediumelectricblue, mark=x] 
	coordinates{
	(1, 44.8)
	(2, 44.7)
	(3, 44.8)
	(4, 44.7)
	(5, 44.8)
	(6, 44.7)
	(7, 44.7)
        (8, 44.8)
	};
    \end{groupplot}
\end{tikzpicture}
\caption{\textbf{Evaluation results using eight different prompts.} The first six prompts use different task descriptions of object detection, while the last two prompts employ random category orders. 
These results show that the performance of different prompts is similar, only a $0.1$ AP gap is observed.} 
\label{fig:different_prompts}
\end{figure}
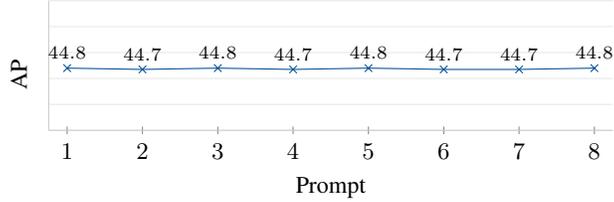
\begin{figure}[t!]
\hsize=\textwidth
\centering
\begin{subfigure}{1.0\textwidth}
    \centering
    \includegraphics[width=0.99\textwidth]{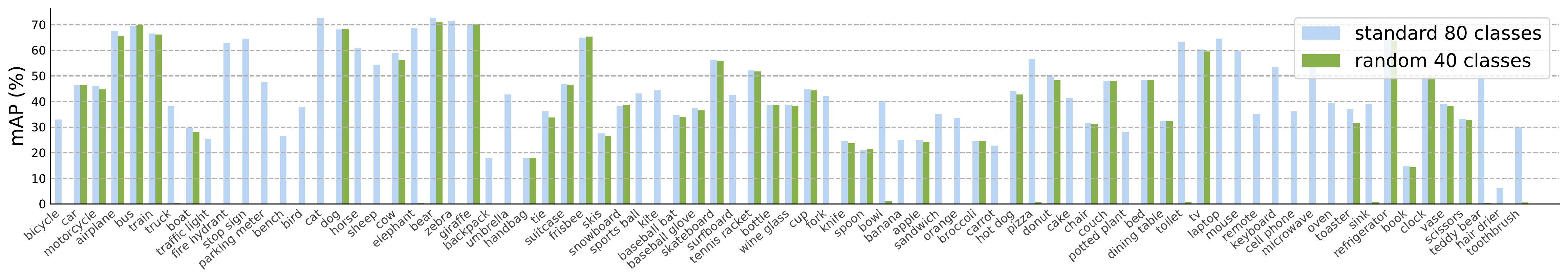}
    \caption{We randomly select 40 classes in random order to form the category set.}
    \label{fig:per_category_40}
\end{subfigure}    
\hspace{0.2in}
\begin{subfigure}{1.0\textwidth}
     \centering
     \includegraphics[width=0.99\textwidth]{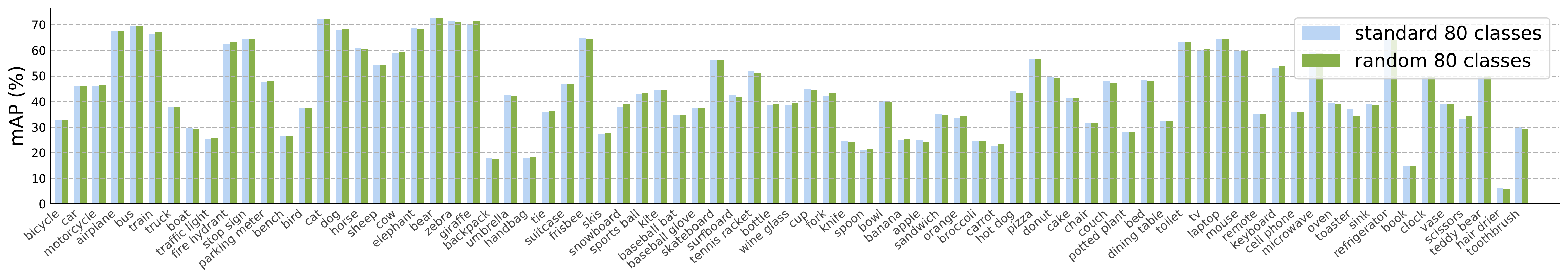}
     \caption{We randomly change the order of 80 classes to form the category set.}
     \label{fig:per_category_80}
\end{subfigure}
\caption{\textbf{Per-category AP on COCO dataset.}
We randomly select some categories to form the category set \texttt{<class>} in language instructions.
}
\label{fig:per_category}
\end{figure}

\textbf{Robustness to Prompt Changes.} Since VisionLLM is trained with random prompts, including random task descriptions and random categories, one may ask whether there is a large performance variance across different prompts. 
To validate the stability of VisionLLM, we conduct experiments using eight different prompts. The first six prompts employ different task descriptions, while the last two prompts involve random category orders. In the case of random category orders, we map the categories back to the COCO standard category order for evaluation.
As shown in Figure \ref{fig:different_prompts}, most evaluation results are distributed closely to $44.8$ AP. The performance differences among prompts are marginal, demonstrating that VisionLLM is robust to different prompts.

\textbf{Instruction Following Capability.} 
As shown in Figure \ref{fig:per_category}, when the prompt only contains $40$ classes, the performance for these categories remains normal, while the performance for the remaining categories is close to zero. This indicates that VisionLLM can dynamically detect objects based on the given class set \texttt{<class>} in instructions while disregarding the other classes that are not mentioned.
This result highlights the flexibility of VisionLLM in adhering to instructions.

\begin{figure}[t!]
    \centering
    \includegraphics[width=1\textwidth]{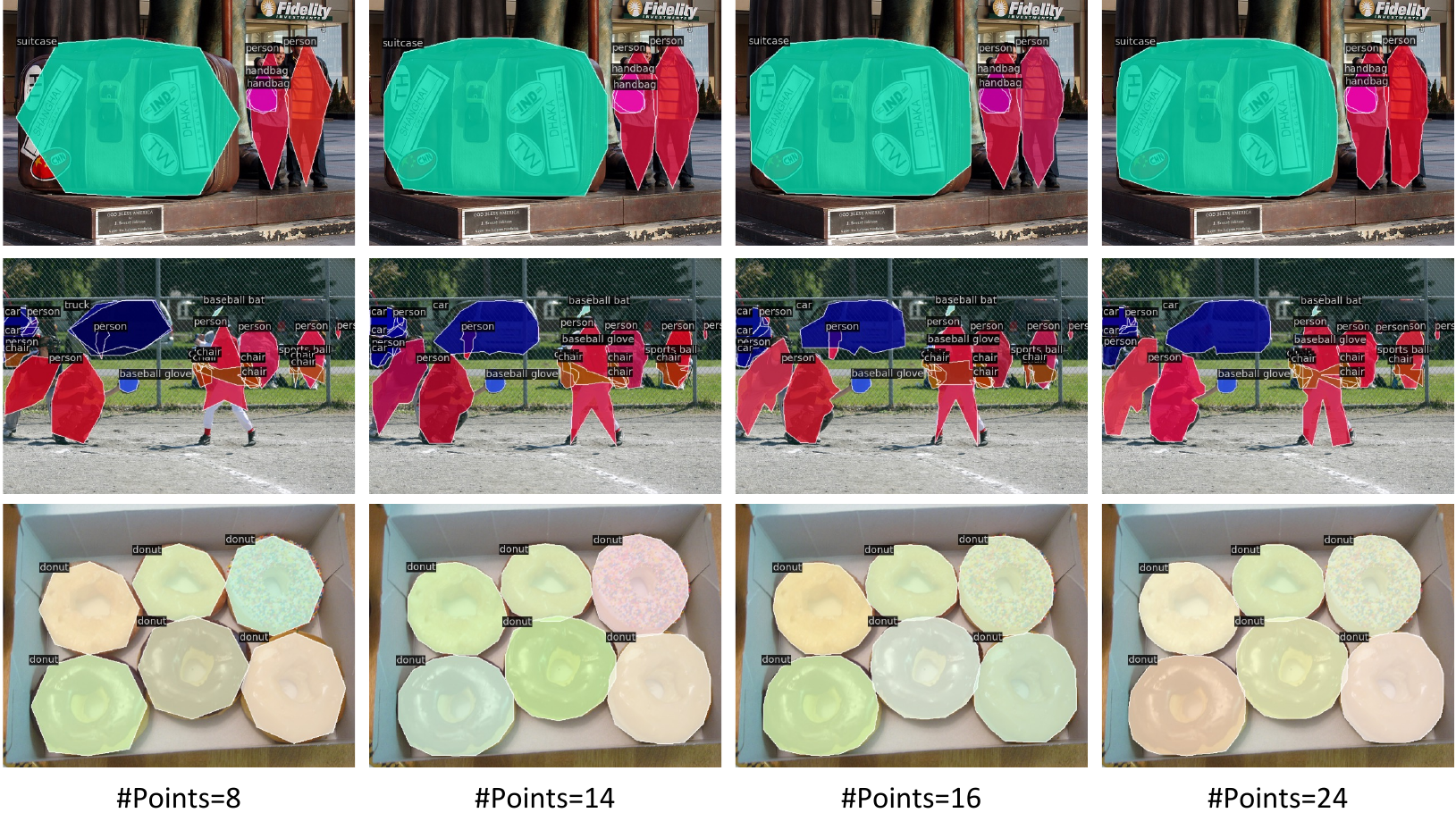}
    \caption{\textbf{Customization of instance masks using the different number of points.} Notably, we only modify the output format mentioned in the prompt, \ie~the number of segmentation points. For more details, please see the example prompts provided in Sec. \ref{sec:ins_seg_prompt}. 
    } 
    \label{fig:vis_n_points}
\end{figure}

\textbf{Output-Format-As-Query \emph{vs.} Seq2Seq.} In VisionLLM, we introduce the output-format-as-query framework for LLM decoder. Alternatively, we also experiment with the sequence generation method like Pix2Seq~\cite{chen2021pix2seq} for object detection with random task descriptions and object categories. However, we find that the loss is hard to converge in this paradigm, which indicates that the seq2seq decoding may need a more detailed design or a longer training schedule for the open-ended visual tasks, while the proposed output-format-as-query framework is more effective for open-ended tasks.

\textbf{Large-Vocabulary Object Recognization.} 
To validate the capacity of VisionLLM in the large-vocabulary scenario, we further conduct the experiments on the challenging dataset LVIS~\cite{gupta2019lvis} with $1203$ categories. Due to the limited number of language tokens, we randomly select 80 classes for training in each iteration. During inference, we divide the $1203$ categories into 16 groups and predict the results in a sliding-window manner. As shown in Table \ref{tab:lvis}, without tricks like federal loss, VisionLLM-R50 can achieve $18.9$ mAP on LVIS.

\section{Qualitative Analysis}

\textbf{Customization of Segmentation Points.} 
In this experiment, we focus on \emph{modifying the output format mentioned in the prompt}, specifically the number of points for instance segmentation (see Sec.~\ref{sec:ins_seg_prompt}). 
The results are visualized in Figure \ref{fig:vis_n_points}.
Remarkably, by increasing the number of points for segmentation, we observe that the model successfully predicts more refined object masks.
This validates the capability of our method to precisely customize the output format, showcasing fine-grained control over the segmentation process.

\begin{figure}[t!]
    \centering
    \includegraphics[width=0.95\textwidth]{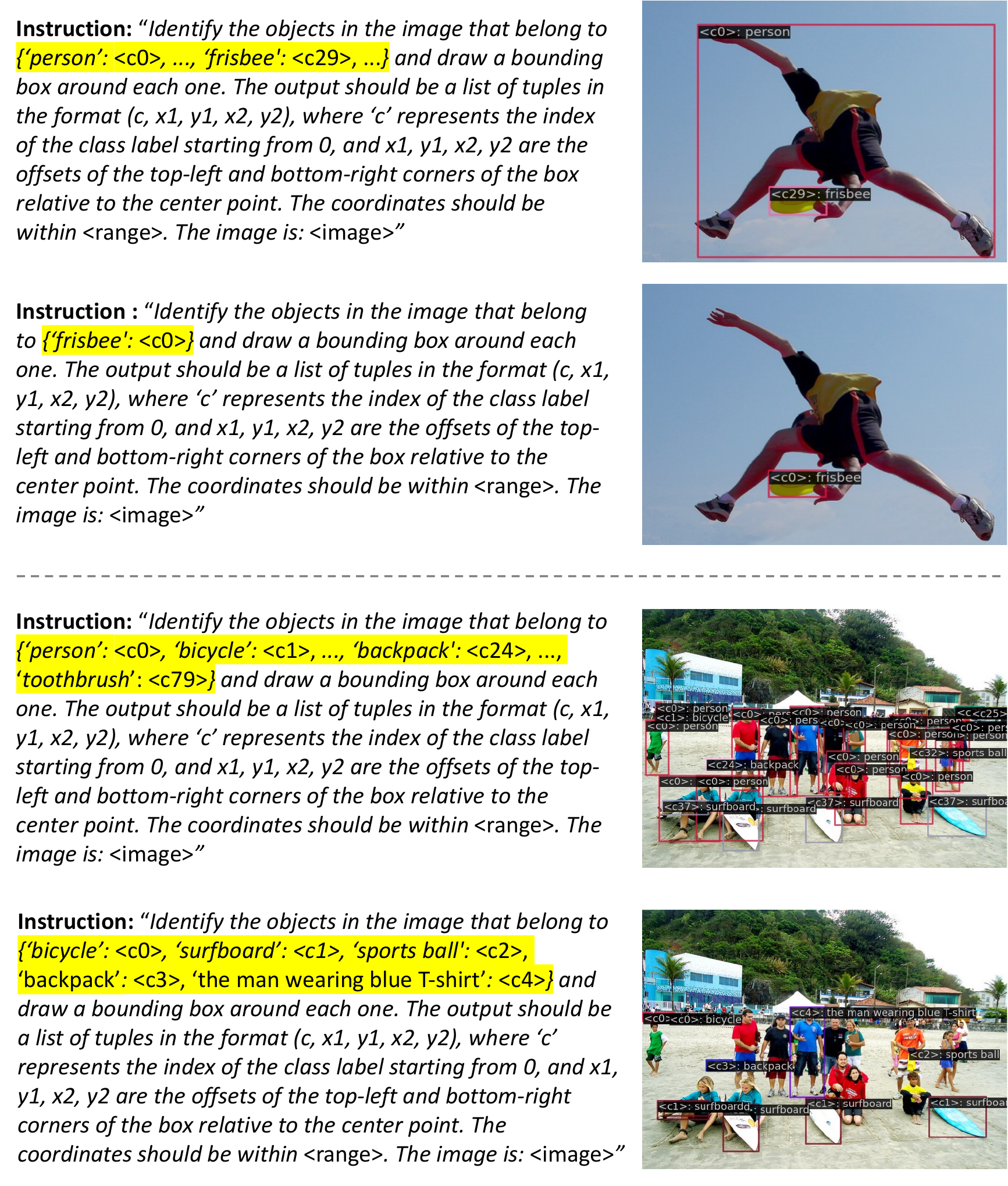}
    \caption{\textbf{Customization of category set.}
    These results demonstrate that VisionLLM can correctly detect the specified objects according to the category set \texttt{<class>} provided in language instructions.
    } 
    \label{fig:vis_cat_set}
\end{figure}

\textbf{Customization of Category Set.}
We change the content of the category set \texttt{<class>} in language instructions and visualize the predictions in Figure \ref{fig:vis_cat_set}. It is observed that the model can correctly predict the object category according to the provided category set. This demonstrates that VisionLLM has a strong capacity to understand and reason over the semantic information of language instructions, which allows it for flexible category customization in open-vocabulary scenarios.

\begin{figure}[t!]
\hsize=\textwidth
\centering
\begin{subfigure}{1.0\textwidth}
    \centering
    \includegraphics[width=0.95\textwidth]{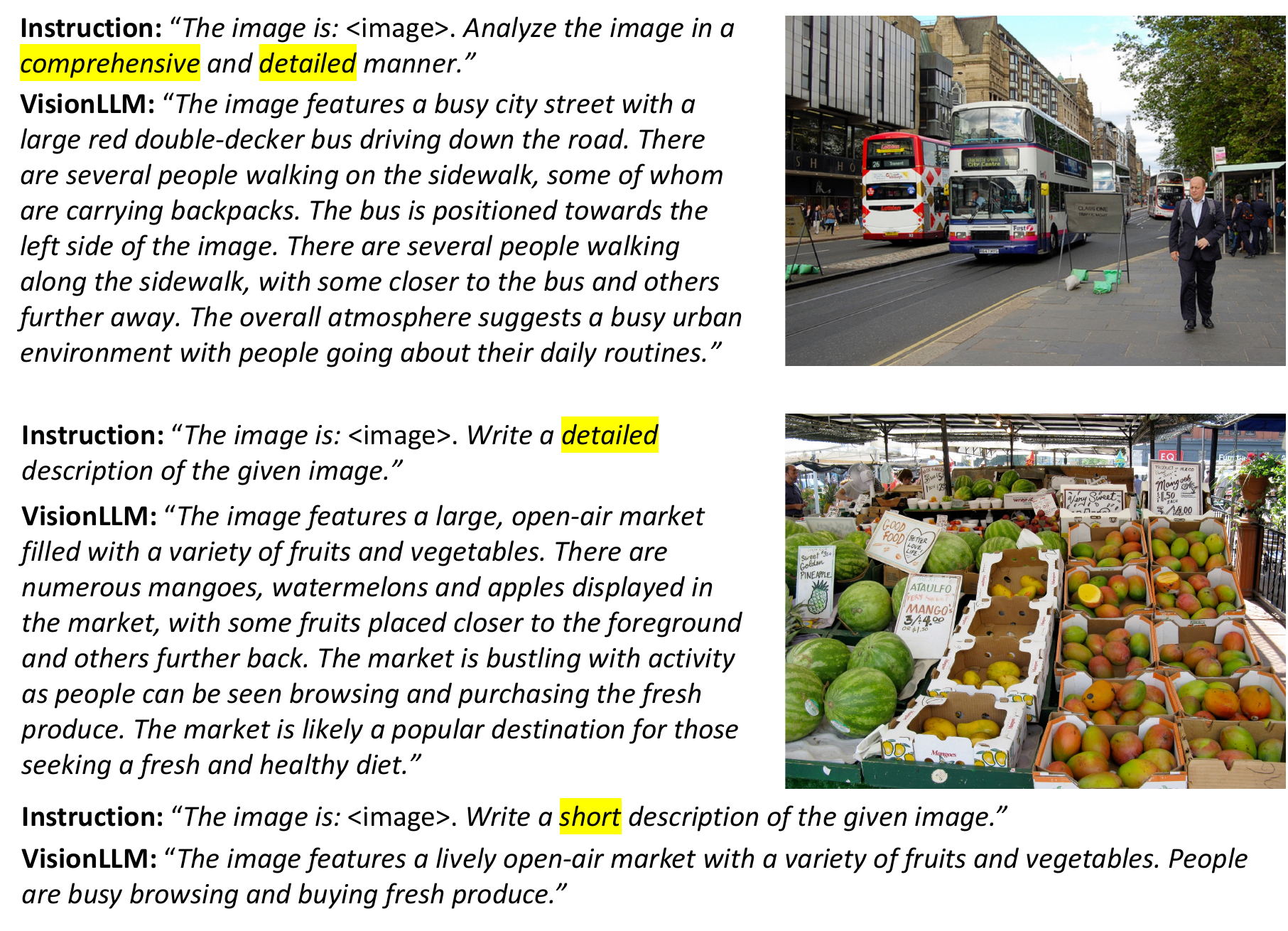}
    \caption{Image description with controllable text length}
    \label{fig:long_description}
\end{subfigure}    
\hspace{0.2in}
\begin{subfigure}{1.0\textwidth}
     \centering
     \includegraphics[width=0.95\textwidth]{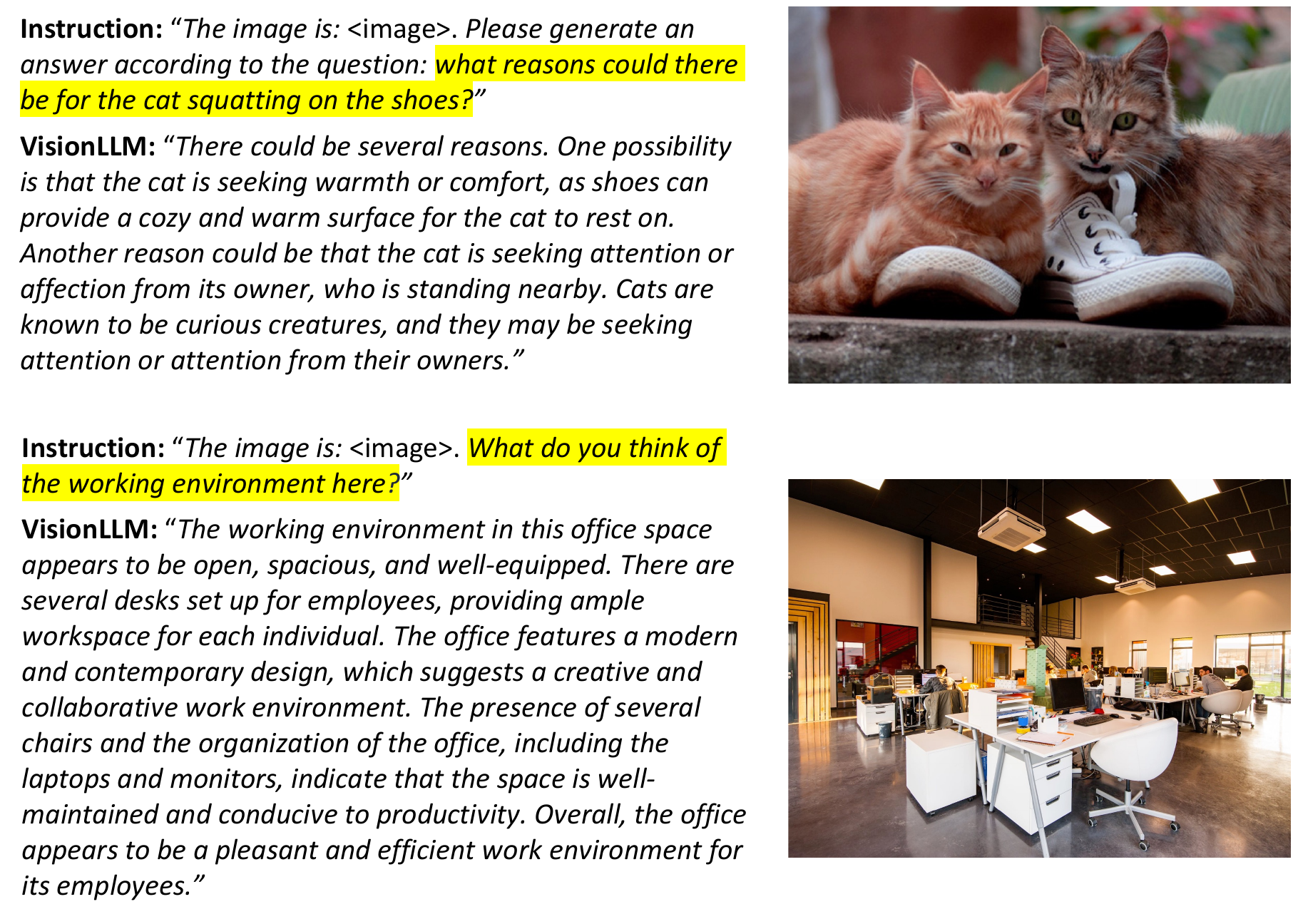}
     \caption{Visual question answering with reasoning}
\end{subfigure}
\caption{\textbf{Visualization of the image description and VQA capabilities of VisionLLM.}
}
\label{fig:vis_vqa}
\end{figure}

\textbf{Image Description \& VQA.} 
Benefiting from the power of LLMs, VisionLLM exhibits a strong ability in generating long descriptions for images and answering visual questions with complex reasoning. We show the examples in Figure \ref{fig:vis_vqa}.

\clearpage

{
\small
\bibliographystyle{plain}
\bibliography{egbib}
}

\end{document}